\documentclass[11pt]{amsart}
\usepackage[margin=1in]{geometry}       
\usepackage[english]{babel}             
\usepackage[utf8]{inputenc}             
\usepackage{amsmath}                    
\usepackage{graphicx}		            
\usepackage[style=alphabetic]{biblatex} 
\usepackage{caption}                    
\usepackage{csquotes}                   
\usepackage[hidelinks]{hyperref}        
\usepackage[all]{hypcap}                
\usepackage{longtable,array}            
\usepackage{booktabs}                   
\usepackage{todonotes}
\usepackage[pagewise, modulo]{lineno}
\title{Modeling Supply and Demand in Public Transportation Systems}

\author{Hala Nelson$^{1*}$}
\author{Miranda Bihler$^2$}
\author{Erin Okey$^{3}$}
\author{Noe Reyes Rivas$^4$}
\author{John Webb$^{1*}$}
\author{Anna White$^5$}
\thanks{*Corresponding authors' email addresses: nelsonhy@jmu.edu, webbjj@jmu.edu\\
$^1$James Madison University, Harrisonburg, VA, USA \\
$^2$Stetson University, DeLand, FL, USA \\
$^3$McMaster University, Hamilton, ON, Canada \\
$^4$Brown University, Providence, RI, USA\\
$^5$East Tennessee State University, Johnson City, TN, USA}

\date{\today}

\newcommand{\acs}{U.S. Census Bureau ACS}

\begin{document}
\maketitle
\thispagestyle{empty}

\begin{abstract}
    We propose two neural network based and data-driven supply and demand models to analyze the efficiency, identify service gaps, and determine the significant predictors of demand, in the bus system for the Department of Public Transportation (HDPT) in Harrisonburg City, Virginia, which is the home to James Madison University (JMU). The supply and demand models, one temporal and one spatial, take many variables into account, including the demographic data surrounding the bus stops, the metrics that the HDPT reports to the federal government, and the drastic change in population between when JMU is on or off session. These direct and data-driven models to quantify supply and demand and identify service gaps can generalize to other cities' bus systems.
\end{abstract}


\section{Introduction}

Harrisonburg City is a small, transit intensive college city, located in the Shenandoah Valley of Virginia. The Harrisonburg Department of Public Transportation (HDPT), which provides bus service throughout the city for local residents and college students, desires to incorporate data driven decision making into its operation. Its goals are to increase ridership, improve the efficiency of their transit system, and enhance its service to the city’s most vulnerable citizens.  The amount of federal funding that the HDPT receives is directly contingent on ridership and efficiency statistics, so improvement in these values can lead to expanded service for Harrisonburg residents.  During the academic year, the HDPT provides shuttles for the campus of James Madison University (JMU), located in the center of Harrisonburg, and off-campus student housing complexes.  For many permanent residents and students with limited personal transportation options, buses are a primary means of travel to allow them to reach areas of the city that are otherwise inaccessible.

 \subsection{Our Contribution} 

In this paper, we propose two new artificial neural network-based supply and demand models for Harrisonburg's bus system using both temporal and spatial data. Our models predict the effect supply changes have on ridership, or demand. Determining how changes in supply affect ridership provides valuable insights into the appropriate allocation of resources. This city-specific analysis of Harrisonburg's transit system, based on the principles and assumptions of supply and demand, provides a detailed path for data-driven decision making into departments of public transportation and other transit systems.

In addition to the temporal and spatial supply and demand models, we propose a new method to identify and quantify service gaps in the city. Resolving service gaps increases the efficiency of the transportation system which can lead to meeting Small Transit Intensive Cities goals and increased funding from the U.S. Department of Transportation. 

Our temporal and spatial supply and demand models take into account both ridership and demographic data. Among the hundreds of variables that factor into the national Social Vulnerability Index (SVI), we incorporate in our models only those that correspond to transportation vulnerability, denoted as transit vulnerability variables (TVV). These variables highlight the vulnerable populations in Harrisonburg who would benefit the most from an optimized transit system. They also provide a better insight into the demographic factors that impact demand for public transportation.

Finally, we propose a method to determine the most significant predictors of supply and demand for artificial neural network based models and apply it to our setting. Popular methods for determining significant predictors were build for traditional statistical models, but we are not aware of literature on how to determine the most significant predictors fo highly nonlinear models such as artificial neural networks, despite the dramatic increase in their use.

 \subsection{Literature Review}
 Previous studies aim to model demand in transit systems. Some use machine learning and artificial neural networks to predict demand as well. The following is an overview of these studies and how they relate to our work:

 \begin{itemize}

\item \cite{currie2004gap} sought to identify geographical and time gaps in transit service of Hobart, Australia. The authors defined a service gap as a location or time-of-day where transit needs are high, while transit service is ``poor or nonexistent.'' To gauge transit service, they created a network supply model of the city and they quantified transit demand using a hand-crafted needs score. In contrast, our supply and demand models are data driven and not hand-crafted, and we quantify service gaps more precisely.  

\item \cite{doi:10.3141/2145-01} is similar to our study in the sense that it directly models transit demand (in Southern California) by stop or station ridership, as opposed to traditional four-step travel demand modeling for corridor- and station-level analyses. Therefore, the bus stop or rail station becomes the unit of analysis. Traditional four-step modeling indirectly estimate transit ridership by first generating vehicle trips, distributing them among origins and destinations, and apportioning travel flows by mode. Direct models estimate ridership at a stop or station on the basis of the intensity of services flowing into it, such as the frequency of buses, properties of the  surrounding environment, such as population densities and demographic attributes, and the stop or station site attributes, such as the presence of a bus shelter or whether a bus stop is marked. Because the direct ridership model focuses on bus stops and their surroundings, it is particularly favorable for estimating the ridership  of transit systems. The authors use a multi-regression model to fit their data. In contrast, we employ an artificial neural network (which fit better than multi-regression and other traditional machine learning models). Our study has the same modeling philosophy and the included attributes as in the referenced paper, but we go further, directly modeling both supply and demand over time and for any stop from the data, and assessing service gaps. 

\item In \cite{DeepLearning2016}, the authors employ an artificial neural network to model demand for bus transit on a stop basis and on a stop-to-stop basis in the dense and crowded city of Seoul. The neural network model utilized data from transit smart cards, and captures the complexities of demand in the city's bus transit system, with its 611 bus lines and 14,287 bus stops, more faithfully than traditional machine learning methods. Our study is similar in the sense that a neural network was the best performer to predict demand, however we do more to predict supply as well, pinpoint under-serviced areas, and assess service gaps. This is all helpful for transit planning and policy making. We also include transit vulnerability variables and assess their impact on supply and demand.

\item \cite{Yan2020}, \cite{9066970}, \cite{https://doi.org/10.1049/itr2.12073}, and \cite{10.1007/978-3-030-94751-4_25} all use machine learning modules to predict demand for ride or bike sharing services. \cite{Yan2020} employs random forests to make predictions, and include variables such as trip-level ride sourcing data, transit supply data, along with publicly available socioeconomic and demographic data. The authors of \cite{9066970} use its machine learning model to forecast bike-sharing supply and demand in Shanghai and incorporate hourly weather conditions into their model.  \cite{10.1007/978-3-030-94751-4_25} creates similar models for Seoul that also include demographic information about the area around bike rental stations and daily COVID-19 case counts.  In contrast to these studies, we model both supply and demand for bus transit systems based on bus stop data and the city's demographic data, and we quantify locational service gaps. We also employ both spatial and temporal data, and settle on artificial neural networks as the best machine learning models to predict supply and demand. 
 
 \item \cite{diab2020rise} analyzed various transit agencies within the Canadian Urban Transit Association. The goal was to identify why transit ridership has leveled-off in Canada despite efforts to increase it. The authors considered ridership among various transit agencies over time, incorporating data from several providers of different sizes. In contrast, we focus solely on a single department of transportation, enabling a closer examination of factors that impact transit and accounting for the distinctive characteristics of Harrisonburg. Our approach in modeling and predicting ridership is also different.


\item While our models focus on the effects of demographics on public transit supply and demand and thus focus on longer time scales, either monthly or yearly, there are examples, such as \cite{8317939} and \cite{9091900}, that use data from mobile phones and transit smart cards to estimate demand for individual trips from one specific location to another over intervals of time as short as every 15 mintues. These are useful studies for adjusting supply on the fly for large and more complex transportation systems that are prone to sudden overloads, however such studies do not integrate demographic data.  Their short time interval approach can be integrated into our work but that is not in the scope of this paper. 

\item Similar to our paper, there are other studies that utilize machine learning to model supply and demand in other sectors than transit systems. These include \cite{HWANGBO2019353} and the references therein.

\end{itemize}

\subsection{Paper Structure}
In Section \ref{sec:overiew_models} we give background information on our models and detail our data sourcing process. Section \ref{sec:details_temporal_models} details the temporal supply and demand models, which analyze the HDPT bus system over the entire city on a month-to-month basis over five fiscal years (2017-2022).  Section \ref{sec:details_spatial_models} details the spatial supply and demand models which focus on the number of routes ran (quantifying supply) and the total ridership (quantifying demand) for each individual stop on city bus routes over a one year period (fiscal year 2019).  In the \hyperref[sec:discussion]{final section} we further discuss our models, make recommendations on how to improve data quality and data collection to increase the accuracy and usefulness of the models, and describe possible directions for future work. The tables in the \hyperref[tab:supply_variables]{appendices} list the variables that we used in our models, their definitions, and their sources.
\section{Overview of Supply and Demand in Bus Transit Systems and Data Wrangling} \label{sec:overiew_models}

This paper focuses on a direct data driven approach to model supply, demand, and the relationships between them. In our context of the bus transit system, supply is related to the HDPT and the services they provide, while demand comes from the needs of the people of Harrisonburg.

Supply directly affects demand, as the amount of buses and routes available contributes to how many people ride the bus; on the other hand, demand does not directly predict supply. The HDPT can modify supply by changing bus schedules, creating new routes, and altering the number of stops, based on the projected needs of riders. However, fluctuations in ridership do not automatically change services without the HDPT's intervention. Thus the quantity supplied at a given point in time can be considered constant.  

An important aspect of our paper is drawing the connection between social vulnerability and transit vulnerability, and incorporating this data into our models. Social vulnerability is traditionally defined as numerous factors that weaken a community's ability when faced with human suffering and financial distress, typically in the face of a natural disaster, as defined by the CDC. From the variables that traditionally impact social vulnerability, we extract those that we assume would impact a community's access to transit services, which we collectively refer to as the transit vulnerability variables (TVV). 

We also quantify supply, demand, and service gaps so that the HDPT can use these quantities to inform its decisions and business models. When transit vulnerability is low and people are able to be self-reliant for their transportation needs, then the quantity demanded for transit is low as well. In contrast, when transit vulnerability is high, the quantity demanded for transit is also high, so we assume that the relationship between transit vulnerability and quantity demanded has a positive slope as demonstrated in Figure \ref{fig:bus_supply_demand}. Since supply, on the other hand, is controlled by the HDPT and does not immediately change in relation to transit vulnerability, it is represented as a horizontal line. 
 Figure \ref{fig:bus_supply_demand} illustrates that an increase in supply allows for a greater quantity of demand to be met, addressing the needs of a more transit vulnerable population.

Figure \ref{fig:bus_supply_demand_gaps} provides additional insight into the relationship between supply and demand in the bus transit system. The difference in quantity supplied and quantity demanded at a constant transit vulnerability is highlighted in gray. This is considered a gap in service, an area or time where transit need is unmet by transit supply. More specifically, it displays a supply shortage, as quantity demanded is greater than quantity supplied.

\begin{minipage}[t]{0.45\textwidth}
    \vspace{-\topskip}
    \includegraphics[width=\textwidth]{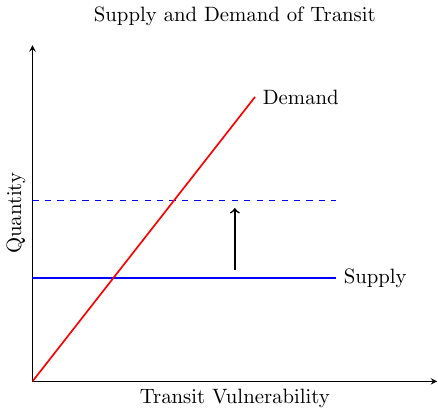}
    \captionof{figure}{Supply remains constant while demand increases according to the transit vulnerability for people within Harrisonburg; an increase in supply allows for a greater quantity of demand to be satisfied.}
    \label{fig:bus_supply_demand}
\end{minipage}
\hfill
\begin{minipage}[t]{0.51\textwidth}
    \vspace{-\topskip}
    \includegraphics[width=\textwidth]{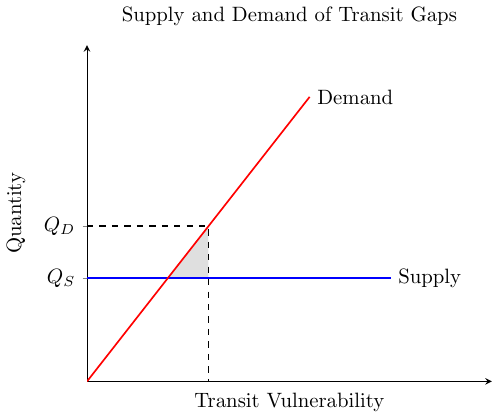}
    \captionof{figure}{At a certain transit vulnerability level, if the quantity demanded is greater than quantity supplied, a service gap exists.}
    \label{fig:bus_supply_demand_gaps}
\end{minipage}
\subsection{Temporal and Spatial Models}

In the temporal model, we examine ridership aggregated
over the entire city, month-by-month, over the period of 5 years, in terms of two supply variables, \textit{vehicle revenue hours} and
\textit{vehicle revenue miles}. We will explain the meanings of these variables shortly, as they are standard
measures for supply in the transportation ecosystem.

In the spatial model, we look at the ridership of each individual stop, over the period of one year, in terms of the number of times a bus passed through it. The space variable in this model refers to the location of a stop.   

We consider the following in our modeling process:
\begin{enumerate}
    \item We assume that supply is only shifted when the HDPT makes a change.
    \begin{quote}
        Although it is understood that many factors contribute to the shifting of a supply curve (increased production costs, government policies, technological innovation, etc.), we assume that the HDPT oversees all of these factors and therefore has the overall control and decision-making power to make changes to their supply.
    \end{quote}
    \medskip
    \item In the temporal model, the population changes during the months that JMU is out of session.
    \begin{quote}
        In Harrisonburg, JMU students make up a large portion of the overall city population, with average annual enrollment being about 20,000 students (roughly 40\% of the city's population). However, all students do not typically remain in the city where their college is located year-round. Thus, the overall population of Harrisonburg must fluctuate based on when JMU is out of session and in session.
     
        To combat the issue of the Harrisonburg population change, we have to approximate the number of students that leave Harrisonburg during the summer and winter breaks. We roughly estimate that 90\% of JMU students that do not attend summer classes would not remain in the city---this takes into account students who may have on-campus jobs, local internships, year-long leases, and any other reasons that may require students to live in Harrisonburg outside of the typical academic year.  This approximation is based on inference and deductive reasoning and is not the result of a methodical study. To improve the accuracy of the population adjustments, we advise to further investigate the true value of students remaining in the location being studied when the university is not in session. 
    \end{quote}
    \medskip
    \item In the spatial model, we assume demographic information does not significantly change between blocks in a given block group.  
    \begin{quote}
        The data we use from the U.S. Census Bureau breaks the demographic information of the city's residents into block groups. To have a more granular form of the data, we calculate the ratio between a block's population to its corresponding block group, then we apply that ratio to each of the variables of transit vulnerability. 
    \end{quote}
\end{enumerate}

\subsection{Data Sourcing and Data Quality}

The data necessary for this project originate from a multitude of sources, which include the United States Census Bureau American Community Survey (ACS), the Office of Institutional Research at JMU, Parking and Transit Services at JMU, the Harrisonburg Department of Public Transportation, and ArcGIS. All the data that we use are either accessible to the public or were collected through a data request to the appropriate department.  Table \ref{tab:data_requests} lists whether the requested data were available or not, and any caveats regarding the original request.

\begin{table}[b]
    \centering
    \begin{tabular}{l r}
        \toprule
        Request & Available \\
        \midrule
        Availability of bird scooters throughout the year & No \\
        HDPT agency profiles & Yes \\
        JMU parking deck spot availability by day from 2017--2022 & No \\
        JMU parking passes sold by semester & Modified; sold annually \\
        JMU student enrollment by semester & Yes \\
        Location of all stops (old and new) by year & Yes \\
        Monthly ridership for transit, paratransit, and JMU & Yes \\
        Number of marked and unmarked stops & No \\
        Paratransit stop data & Yes \\
        Shared rides usage throughout the year (Uber, Lyft, etc.) & No \\
        Transit Vulnerability Variables (TVV) by block for 2017--2022 & Modified; by block group \\
        Social Vulnerability Index (SVI) by census tracts from 2017--2022 & Modified; only 2018 available \\
        STIC (Small Transit Intensive Cities) apportionments & Yes \\
        Transit stop data for all routes & Yes \\
        \bottomrule
    \end{tabular}
    \caption{Data requests and availability of data.}
    \label{tab:data_requests}
\end{table}

\subsubsection{Data Quality} \label{sssec:data_quality}

 There are many instances where the data were not collected in an efficient manner or were disorderly in its original format. This required excessive cleaning and pre-processing. For example:

\begin{itemize}
\item Certain data (e.g., the number of bus drivers employed, the number of vehicles in operation) were not separated by time intervals in a consistent way. 
\item The bus stop data that we received from the HDPT contained many errors, missing values, and duplicated columns.
\item There is an overall lack of efficient data collection and retrieval methods in various departments. For example, to obtain the parking deck spot availability data from JMU over the past five years, the JMU Department of Transit Services would have to manually request the data for each day. Moreover, only a limited number of parking decks track the daily space usage. We decided that the number of parking passes sold by JMU each academic year would be an adequate indicator of private car usage on-campus.

\item The data corresponding to the individual stops and their ridership (boardings, alightings, and the total number of trips separated by stop and stop ID) that we received from the company that managed the HDPT's data were grossly damaged. For example, most stops that were labeled within the inner-campus shuttle route were strewn throughout the city of Harrisonburg and were not on-campus.  When investigating further, we discovered that all of the stops in the data file were mislabeled.  While the data for ridership on each route is correct, we were forced to find alternate sources for data on the ridership for each stop.
\end{itemize}

\subsubsection{Census Data}

The U.S.\ Census Bureau breaks Harrisonburg into 11 census tracts.  These are then further subdivided into a total of 27 block groups, which is the smallest spatial unit that most data are reported on each year.
 Each block group is comprised of individual blocks---Harrisonburg has a total of 631 blocks.  However, only basic population data on blocks from the decennial census are publicly available.  We used the most recent block data available to calculate how the population in each block group was distributed within the individual blocks.  This allowed us to make a more accurate estimate of the population serviced by each individual bus stop based on what blocks were located around it.

\subsubsection{Social Vulnerability}
    
The social vulnerability index (SVI) is a measure used by the Center for Disease Control (CDC) and other government institutions to measure potential adverse effects to humans by external pressures. The index itself is a positive number ranging from 0 to 1, with 0 indicating low social vulnerability and 1 indicating high social vulnerability. According to \cite{atsdr2022CDC},

\begin{quote}
    Every community must prepare for and respond to hazardous events, whether a natural disaster like a tornado or disease outbreak, or a human-made event such as a harmful chemical spill. A number of factors, including poverty, lack of access to transportation, and crowded housing may weaken a community’s ability to prevent human suffering and financial loss in a disaster. These factors are known as social vulnerability.
\end{quote}

The CDC calculates the social vulnerability of a community based on 15 factors, grouped under four themes: 

\begin{minipage}[t]{0.45\textwidth}
\begin{itemize}
    \item Socioeconomic Status
    \begin{itemize}
        \item Below Poverty 
        \item Unemployed 
        \item Income 
        \item No High School Diploma 
    \end{itemize}
    
    \item Household Composition \& Disability
    \begin{itemize}
        \item Age 65 or Older 
        \item Age 17 or Younger 
        \item Older Than Age 5 With a Disability 
        \item Single-Parent Households 
    \end{itemize}
\end{itemize}
\end{minipage}
\begin{minipage}[t]{0.45\textwidth}
\begin{itemize}
    \item Minority Status \& Language
    \begin{itemize}
        \item Minority 
        \item Speaks English ``Less Than Well'' 
    \end{itemize}
    
    \item Housing \& Transportation
    \begin{itemize}
        \item Multiunit Structures 
        \item Mobile Homes 
        \item Crowding 
        \item No Vehicle 
        \item Group Quarters
    \end{itemize}
\end{itemize}
\end{minipage}
\vspace{1em}

\begin{figure}
\includegraphics[width=0.5\textwidth]{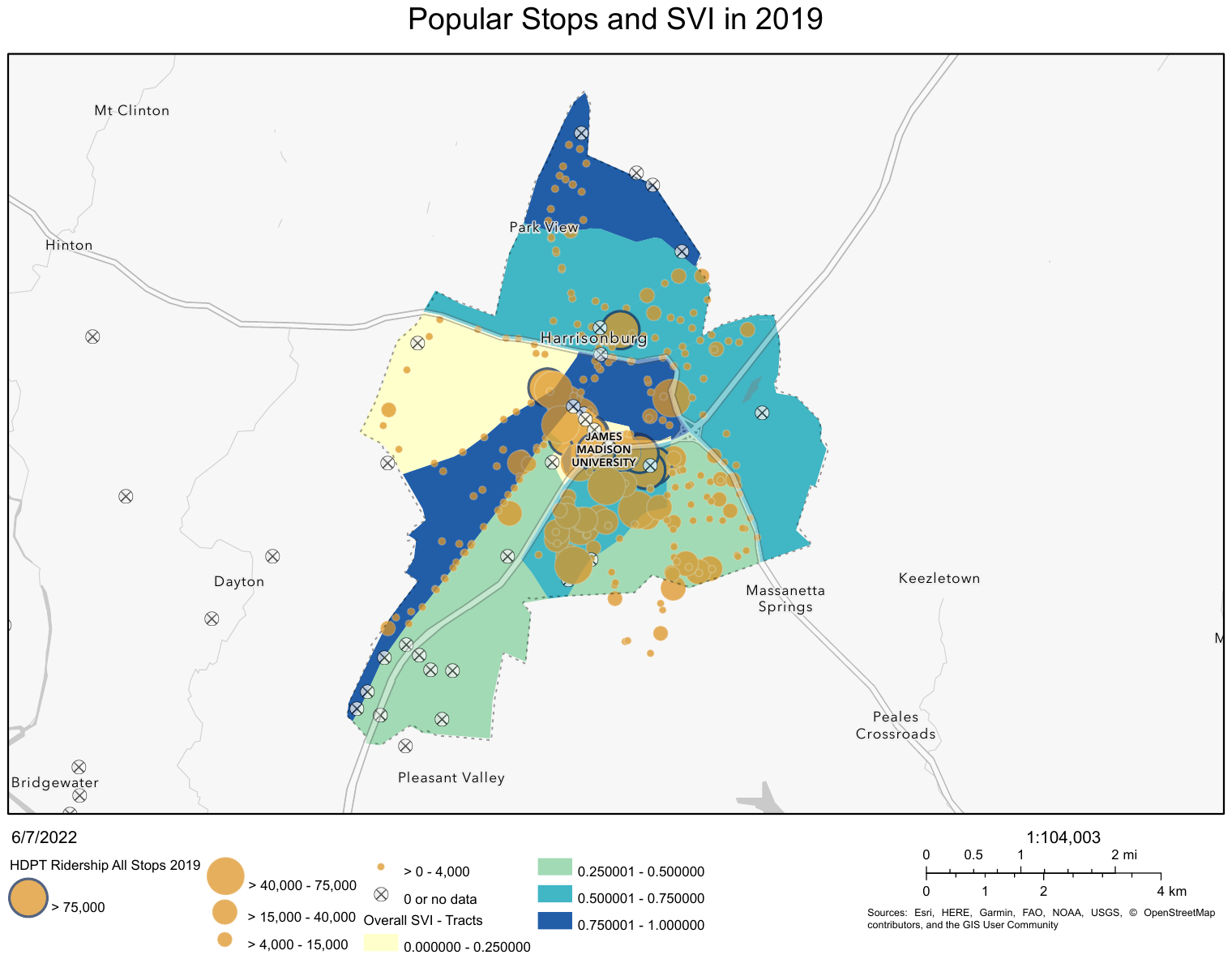}
\caption{Popular bus stops during 2019. Larger circles reflect more boardings and alightings. The colors on the map reflect the social vulnerability index by census track.}
\label{SVI_busstops}
\end{figure}

As of 2018, the city of Harrisonburg has an average SVI of 0.79, indicating a medium to high level of social vulnerability. A more detailed description of SVI per census tract is given in Table \ref{tab:census_tract_svi} and we visually represent this in Figure \ref{fig:svi_map}. The SVI is a helpful tool to understand the city population, however, the data is only available by census tract. Figure \ref{SVI_busstops} shows a map of HDPT bus stops overlaid on a map of SVI by census tract. The size of the circle indicates the popularity of the stop with more popular stops being represented by larger circles. Since we need to assess transit supply and demand by the more granular census block, instead of the census tract, we do not include the SVI index in our models, but extract select factors. We refer to these as the transit vulnerability variables (TVV).

\begin{table}[ht]
	\begin{minipage}{0.5\linewidth}
        \centering
        \begin{tabular}{c c c}
            Census Tract & Population & SVI \\
            \hline
            1.01 & 5015 & 0.7236 \\
            1.02 & 5696 & 0.7282 \\
            2.03 & 1901 & 0.2533 \\
            2.04 & 4430 & 0.7931 \\
            2.05 & 6001 & 0.2650 \\
            2.06 & 4725 & 0.0761 \\
            2.07 & 6231 & 0.7299 \\
            3.01 & 3575 & 0.1122 \\
            3.02 & 5679 & 0.8001 \\
            4.01 & 3577 & 0.8111 \\
            4.02 & 6561 & 0.7432
        \end{tabular}
        \caption{A table of SVI for Harrisonburg, VA by census tract.}
        \label{tab:census_tract_svi}
	\end{minipage}\hfill
	\begin{minipage}{0.45\linewidth}
		\centering
		\includegraphics[width=\textwidth]{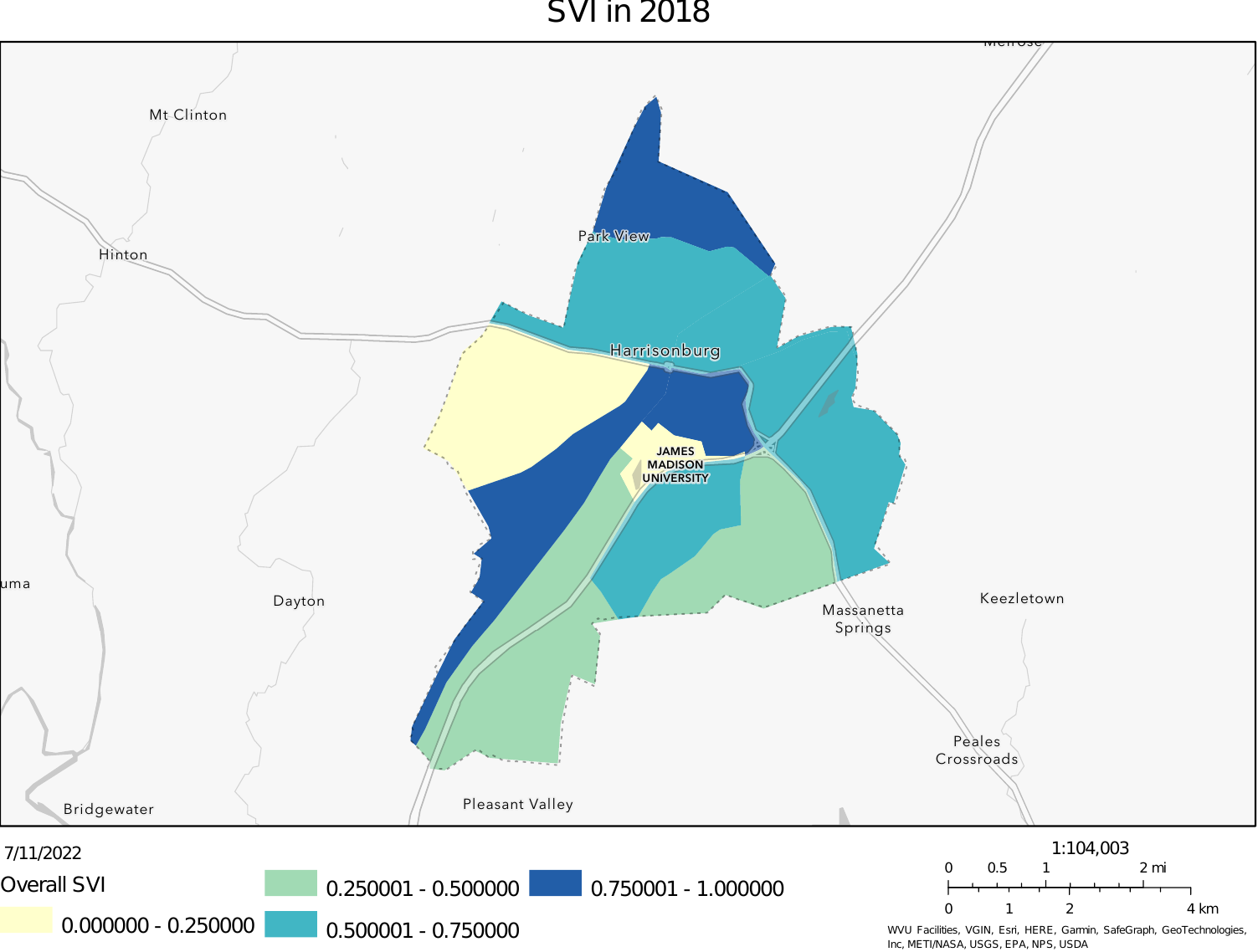}
		\captionof{figure}{A map of SVI for Harrisonburg, VA by census tract.}
		\label{fig:svi_map}
	\end{minipage}
\end{table}

\subsubsection{JMU Parking}
The amount of parking spaces available to JMU students impacts the HDPT's ridership.  Anecdotally, the HDPT executives noticed that ridership numbers on certain routes increased after a new parking deck opened on the eastern edge of campus. Data on the number of available parking spaces on-campus and the daily usage of these parking spaces were not available.  Instead, we were able to obtain information on the number of JMU parking permits in use each semester, which we incorporated into the model.   
\subsection{Data Trends}

We graph the monthly ridership of various types of bus transportation systems in Harrisonburg over the past 20 years. The plots in Figures \ref{fig:transit_ridership} and \ref{fig:jmu_transit_ridership} exhibit similar trends, indicating the large influence of JMU students and faculty on transit ridership. All types of transit experienced a general increase in ridership between 2002 and 2020; ridership decreased in 2020 due to COVID-19, indicated through bolded lines. However, ridership across all transit has since begun to increase and is now approaching pre-COVID-19 levels.

The general trend also indicates that ridership of general transit appears to depend on the JMU population being present. As the figures illustrate, there are significant drops in ridership when JMU is not in session, according to the academic calendar. Not only is ridership impacted by the presence of JMU students, but the general population of Harrisonburg also fluctuates accordingly. This lack of a constant population throughout the year adds complexity to the ridership for the transit system of Harrisonburg, and we address this when constructing the models.

\begin{minipage}{0.47\textwidth}
    \includegraphics[width=\textwidth]{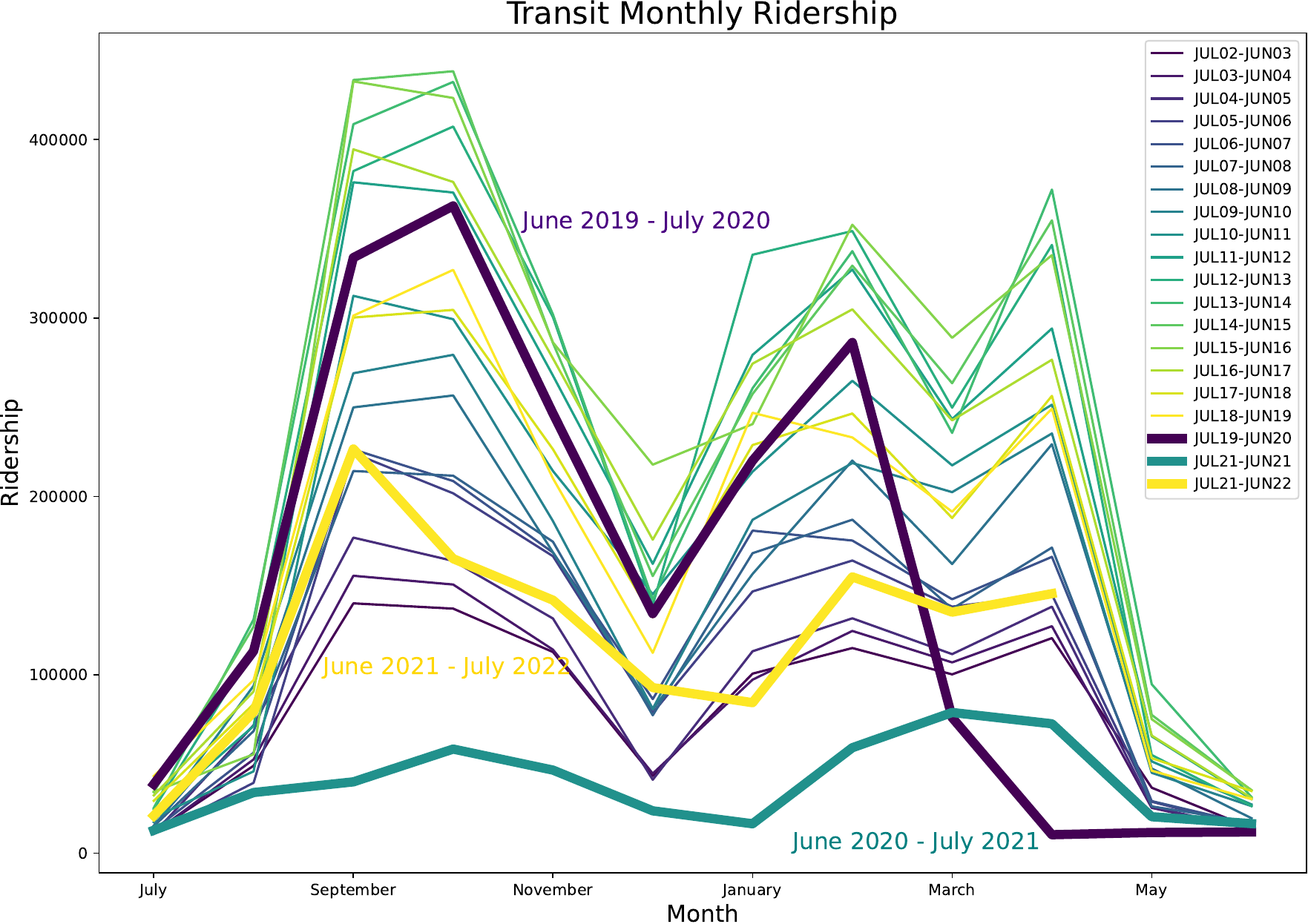}
    \captionof{figure}{Monthly general transit ridership between June 2002 and July 2022.}
    \label{fig:transit_ridership}
\end{minipage}
\hfill
\begin{minipage}{0.47\textwidth}
    \includegraphics[width=\textwidth]{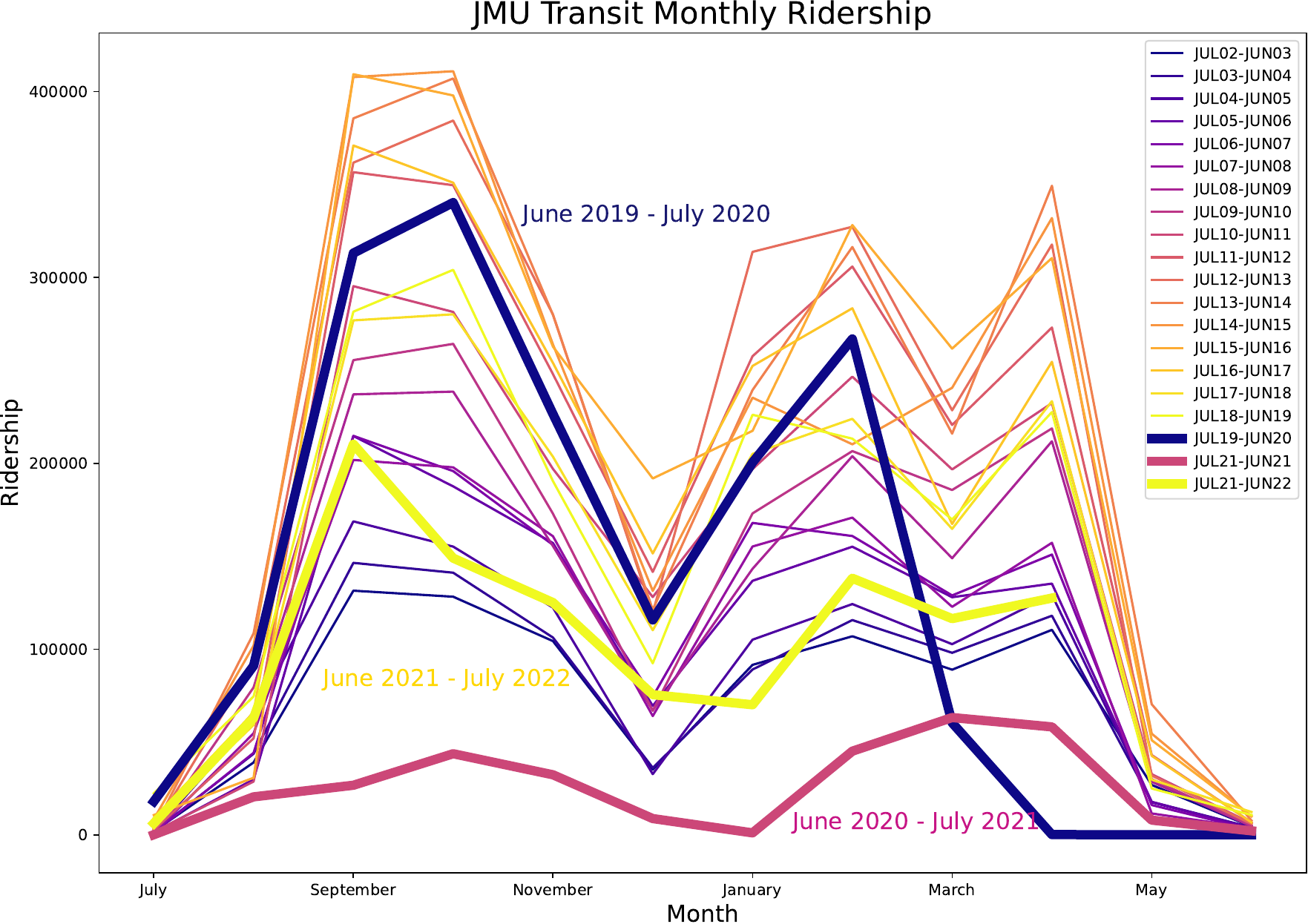}
    \captionof{figure}{Monthly JMU transit between June 2002 and July 2022.}
    \label{fig:jmu_transit_ridership}
\end{minipage}
\subsection{Stops with Significant Service Gaps} \label{ssec:service_gap_real} 

Figures \ref{fig:ridership_vs_routes_outlier} and \ref{fig:ridership_vs_routes} show the number of riders that boarded or alighted at a stop against the number of routes that passed through that stop during 2019. Figure \ref{fig:ridership_vs_routes_outlier} shows all of the stops that were analyzed while Figure \ref{fig:ridership_vs_routes} excludes the stop with highest ridership. Excluding this high-demand stop allows us to use a smaller scale for \textit{Total Riders} so that the remaining stops are visible in more detail.

We see supply shortages in areas that have relatively few routes but high ridership. Conversely, areas with high numbers of routes ran but relatively low ridership have a supply surplus. The color gradient in Figure \ref{fig:ridership_vs_routes}, computed as the total number of riders relative to the total number of bus routes per stop, represents these shortages and surpluses. 

These numbers also generally reflect how many routes a stop is on. However, this is not the case when a stop is passed through multiple times on one route. Harrisonburg Crossing at Walmart, one of these excepted stops, is on two routes, but since one of those routes passes through it twice, it is counted as being on three routes. 

Using real stop data, we can immediately assess service gaps for each existing stop. In section 4, we will use spatial supply and demand models to assess service gaps at any existing or new stop in the city. For these particular stops with service gaps, pinpointed from real data, we can use the spatial supply and demand models to predict the correct level of supply and demand that doesn't result in surplus or under-service. The goal is to minimize service gaps at every location in the city.

Upon investigating these service gaps further, we determine that the HDPT is effectively adjusting supply, overall, especially with the influx of population in Harrisonburg when the academic semester is in session; however there are a number of individual stops with high ridership, particularly on the outskirts of Harrisonburg, where increases in service may better serve the residents.

\begin{minipage}[t]{0.48\textwidth}
    \centering
    \includegraphics[width=\textwidth]{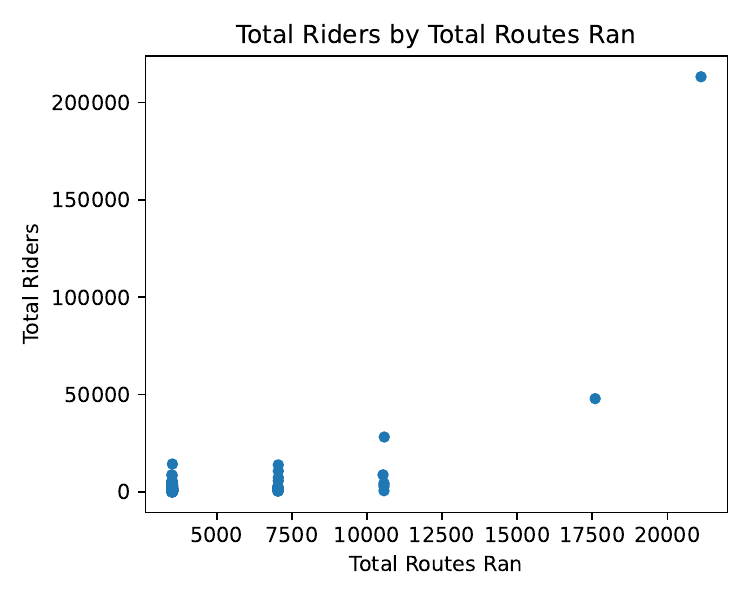}
    \captionof{figure}{Stop ridership against city routes ran to demonstrate which stops have supply surpluses and shortages.}
    \label{fig:ridership_vs_routes_outlier}
\end{minipage}
\hfill
\begin{minipage}[t]{0.48\textwidth}
    \centering
    \includegraphics[width=\textwidth]{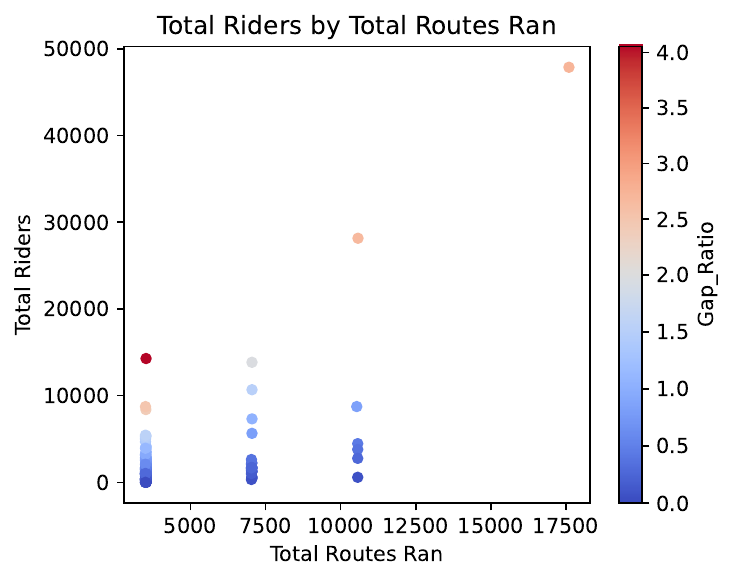}
    \captionof{figure}{City routes ran against stop ridership with an outlier removed. The color gradient represents the ratio between Total Riders and Total Routes Ran ($\text{Total Riders}/\text{Total Routes Ran}$).}
    \label{fig:ridership_vs_routes}
\end{minipage}

\begin{figure}
    \centering
    \includegraphics[width=0.75\textwidth]{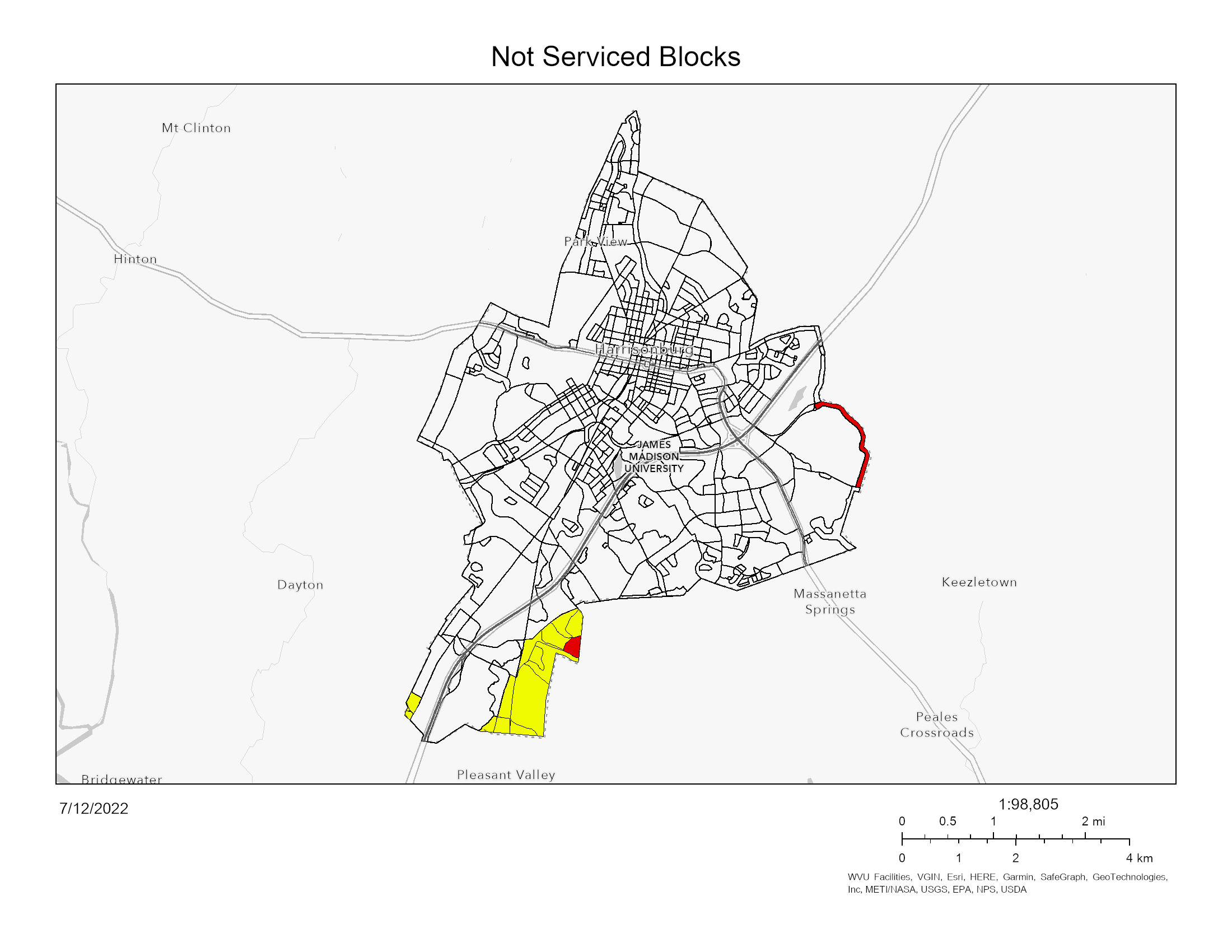}
    \caption{The highlighted red zones are not serviced at the time of this study. The highlighted yellow zones were not serviced during 2019 but this has since been rectified.}
    \label{fig:not_serviced}
\end{figure}

In addition to identifying bus stops with service gaps, we pinpoint completely unserviced areas in Figure \ref{fig:not_serviced}. This is only based on ArcGIS mapping and the given bus stop and bus route data.

\subsection{Overview of Temporal and Spatial Supply and Demand Models} \label{ssec:model_overview}


We employ two different types of models for supply and demand: the first one is a temporal model evaluating monthly changes of supply and demand of the entire city over the period of five fiscal years while the second one is a spatial model examining supply and demand at each individual bus stop within one year. 


Ideally, we would combine spatial and temporal data for a more reflective model looking at supply and demand monthly changes per stop, however that data was either not available or corrupted, as discussed in section \ref{sssec:data_quality}. Our recommendation is for the HDPT to acquire that data as it is crucial for building an accurate predictive model.
\section{Details of Temporal Supply and Demand Models} \label{sec:details_temporal_models}

We use monthly data from July 2017 to June 2022, a total of five fiscal years, to analyze the supply and demand of the HDPT bus system over the entire city. Our two quantifiers of supply are \textit{vehicle revenue hours} and \textit{vehicle revenue miles}. \textit{Vehicle revenue hours} are defined as the hours that vehicles are scheduled to or actually travel while in revenue service and \textit{vehicle revenue miles} use the same definition in the unit of miles. These definitions are set by the Federal Transit Administration in the National Transit Database Glossary. Our quantifier for demand is \textit{number of passenger trips} which demonstrates ridership. 
\subsection{Variables Included in the Supply Model}

To model supply, we use the variables listed in Appendix \ref{tab:supply_variables} as input variables, and the vehicle revenue miles and vehicle revenue hours as target variables. We use vehicle revenue miles and vehicle revenue hours as measures for supply since these same quantifiers are used in the Small Transit Intensive Cities (STIC) apportionments. In the past five years, the HDPT has been within 0.3 vehicle revenue miles per capita from attaining the federal goal set for them. When a new benchmark is achieved, greater funding will be distributed to the city. Predicting vehicle revenue miles and vehicle revenue hours gives the HDPT informed insights on a category that is achievable which, in turn, increases the HDPT budget from the STIC apportionments.



\begin{figure}
    \centering
    \includegraphics[width=\textwidth]{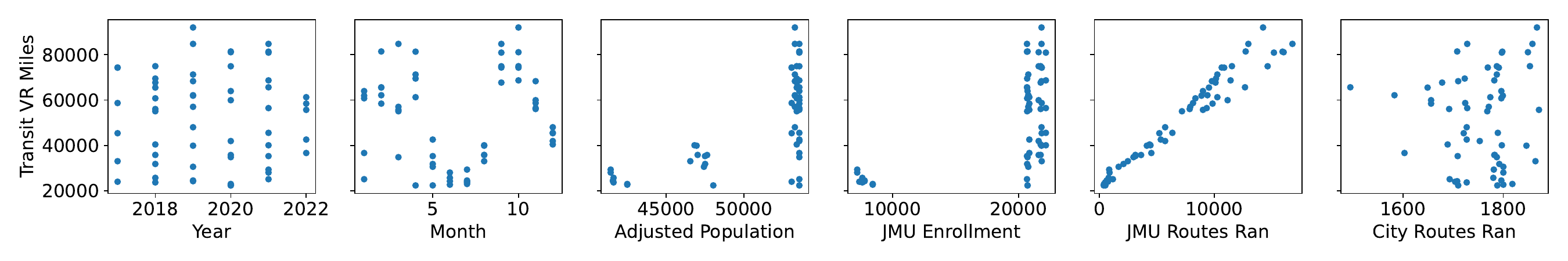}
    \includegraphics[width=\textwidth]{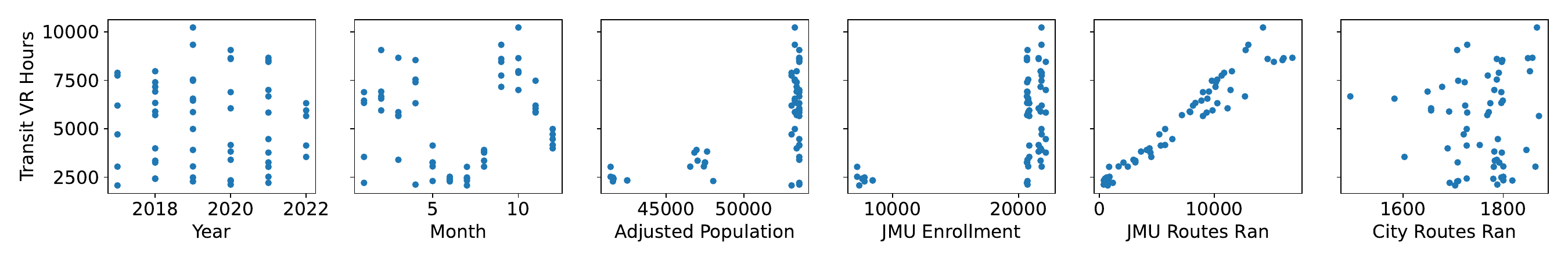}
    \caption{Scatter plots of $RevenueMiles$ and $RevenueHours$ (quantifiers of supply) against each input variable of the supply function.}
    \label{fig:temporal_supply_vs_inputs_plots}
\end{figure}

\noindent We therefore use the following functions to quantify supply:

\begin{equation*}
\begin{aligned}
    &RevenueMiles_{predicted}= Supply_1\left(AdjPop, JMUEnrollment, JMURan, CityRan, t_{year},t_{month}\right) 
    \\&RevenueHours_{predicted}= Supply_2\left(AdjPop, JMUEnrollment, JMURan, CityRan, t_{year},t_{month}\right) 
\end{aligned}
\end{equation*}

\noindent where $AdjPop$ is the adjusted population of the total population in Harrisonburg with JMU's enrollment accounted for; $JMUEnrollment$ is JMU's enrollment numbers, and $JMURan$ and $CityRan$ are the number of bus routes running in a single day on and off campus, respectively. Plots of each input variable against each target variable can be seen in Figure \ref{fig:temporal_supply_vs_inputs_plots}.


\subsection{Variables Included in the Demand Model}
We quantify demand using the number of passenger trips, which is the sum of passenger boardings and alightings. This is the most logical variable to reflect ridership for the transit system and it is one of the priorities of HDPT.

\begin{figure}
    \centering
    \includegraphics[width=\textwidth]{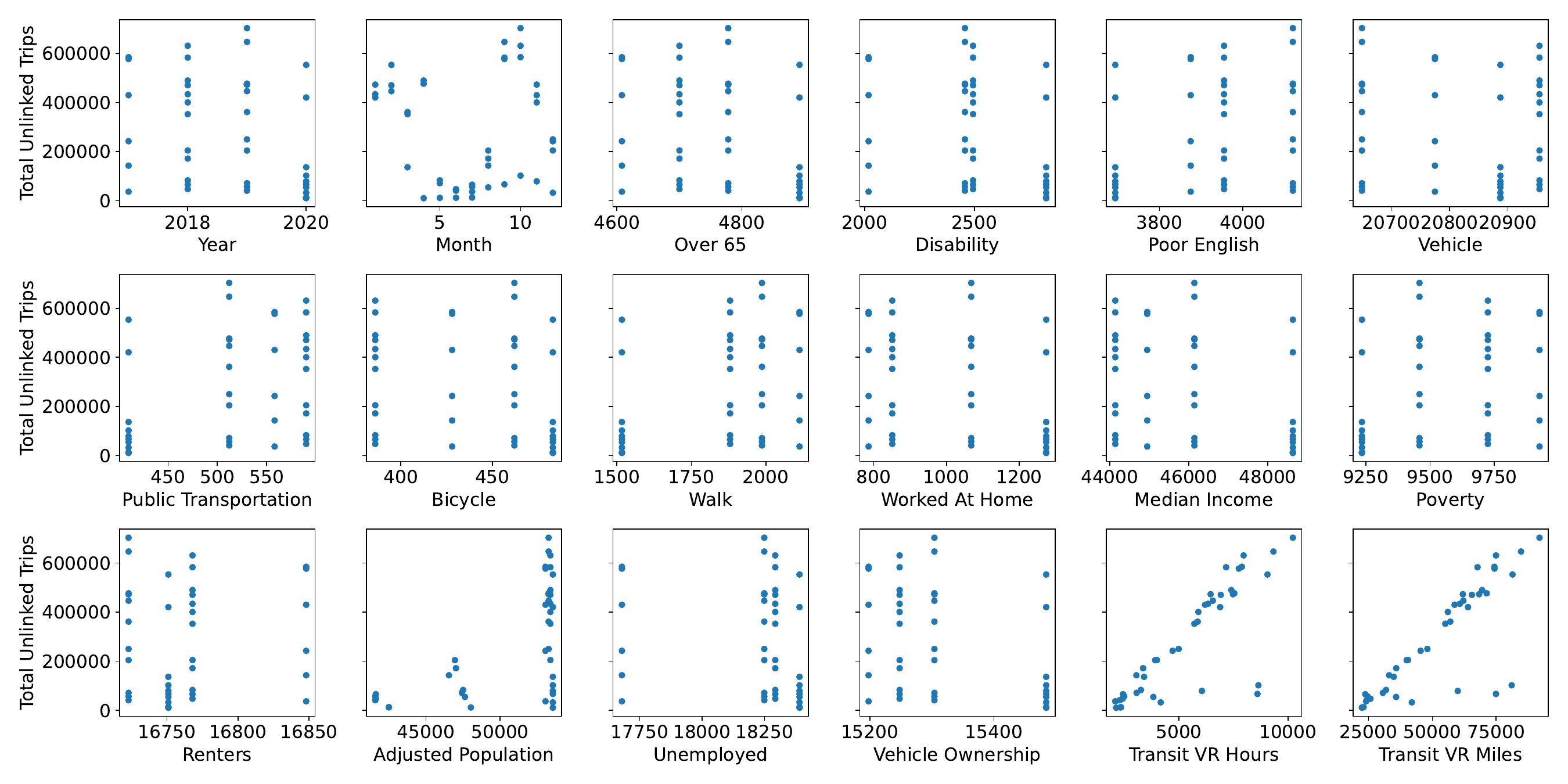}
    \caption{Scatter plots of $NumberPassengerTrips$ (quantifier for demand) against each input variable of the demand function.}
    \label{fig:temporal_demand_vs_inputs_plots}
\end{figure}

\noindent We therefore use the following function to quantify demand:

\begin{align*}
    NumberPassengerTrips_{predicted} =&Demand \big(RevenueMiles_{actual}, RevenueHours_{actual}\\&AdjPop, TVV,t_{year},t_{month}\big)
\end{align*}

\noindent where $RevenueMiles_{actual}$ is the actual vehicle revenue miles and $RevenueHours_{actual}$ is the actual vehicle revenue hours, as provided by the HDPT, $AdjPop$ is Harrisonburg's adjusted population with JMU's enrollment accounted for, $TVV$ are the transit vulnerability variables referred to in section 2.2.3, which include variables such as: population age 65 and over, with disability, below poverty level, speak English ``less than well'', renter population, vehicle ownership, and means of transportation. Plots of each input variable against the target variable, \textit{NumberPassengerTrips}, can be seen in Figure \ref{fig:temporal_demand_vs_inputs_plots}.

Note that we used the actual supply values for the build stage of the temporal demand model, but in deployment and prediction stage, we can use the values predicted by the temporal supply model---that is: 

\begin{align*}
    NumberPassengerTrips_{predicted} =&Demand \big(RevenueMiles_{predicted}, RevenueHours_{predicted}\\&AdjPop, TVV,t_{year},t_{month}\big)
\end{align*}

\subsection{Machine Learning Models and Results}

We experiment with various machine learning models: linear regression, polynomial regression, random forests, and neural nets with different architectures. In each of these models, we used 80\% of the data points as the training set and the remaining 20\% as the test set.

\begin{table}
    \centering
    \begin{tabular}{c c c c c}
    \toprule
    Machine Learning Model & \multicolumn{2}{c}{Root Mean Square Error} & \multicolumn{2}{c}{Relative Root Mean Square Error} \\
    & Miles & Hours & Miles & Hours \\
    \midrule
    Linear Regression & 488.21674025 & 3657.31427638 & 0.09954634 & 0.07651245 \\
    Polynomial & 798.29113149 & 5357.43653502 & 0.16276984 & 0.11207968 \\
    Neural Network & 171.13969417 & 1652.09654682 & 0.03489502 & 0.03456251 \\
    Random Forest & 484.01463476 & 4592.11630587 & 0.09868954 & 0.09606888 \\
    \bottomrule
\end{tabular}
    \caption{Summary of performance measures of machine learning models for temporal supply, quantified as $RevenueMiles$ and $RevenueHours$.}
    \label{tab:supply_error}
\end{table}

\begin{figure}
    \includegraphics[width=0.5\textwidth]{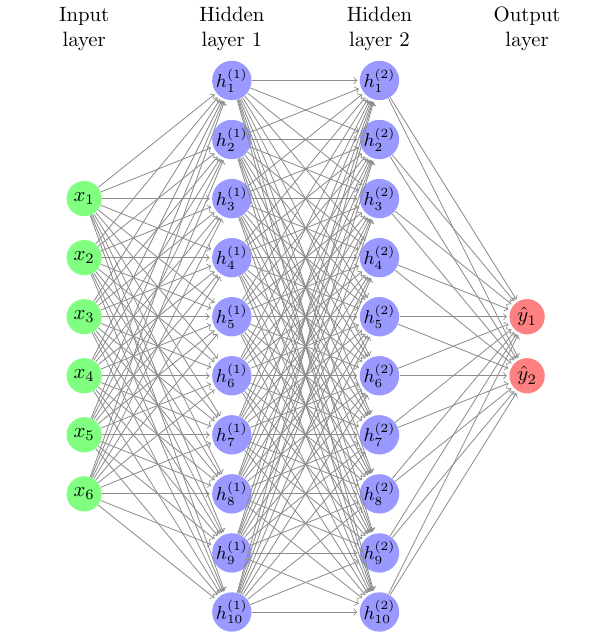}
    \caption{Visual representation of neural network architecture.}
    \label{fig:nn}
\end{figure}

As seen in Table \ref{tab:supply_error}, neural nets performed the best in predicting supply, having the lowest relative root mean squared error. This is a feed forward fully connected neural network that uses ReLU activation function. We experimented with a few architectures, and employed a network with two hidden layers, with ten nodes each (see Figure \ref{fig:nn}).

\begin{figure}
    \centering
    \includegraphics[width=0.45\textwidth]{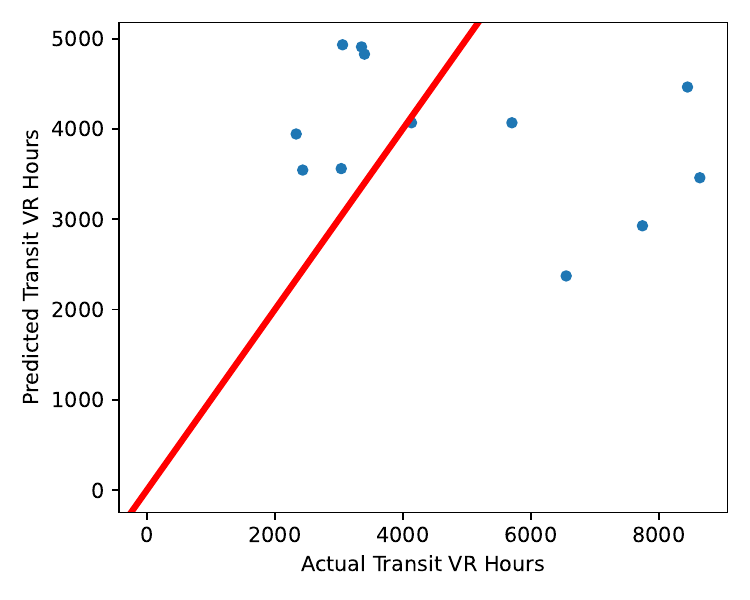}
    \includegraphics[width=0.45\textwidth]{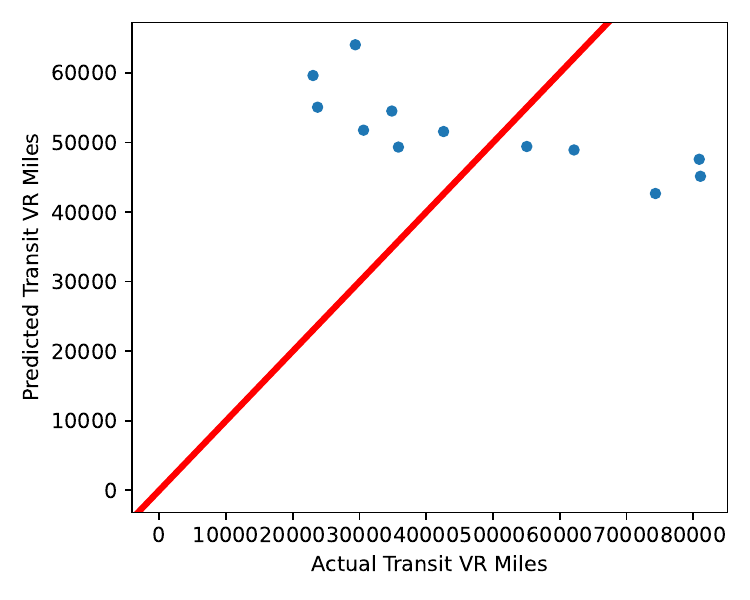}
    \caption{Each point represents the predicted supply versus the actual supply of the test dataset for a specific month, measured using $RevenueHours$ on the left and $RevenueMiles$ on the right. Points close to the red line of equation $y=x$ are the months where prediction match real data.}
    \label{fig:temporal_supply_actual_vs_predicted}
\end{figure}

\begin{figure}
    \centering
    \includegraphics[width=0.45\textwidth]{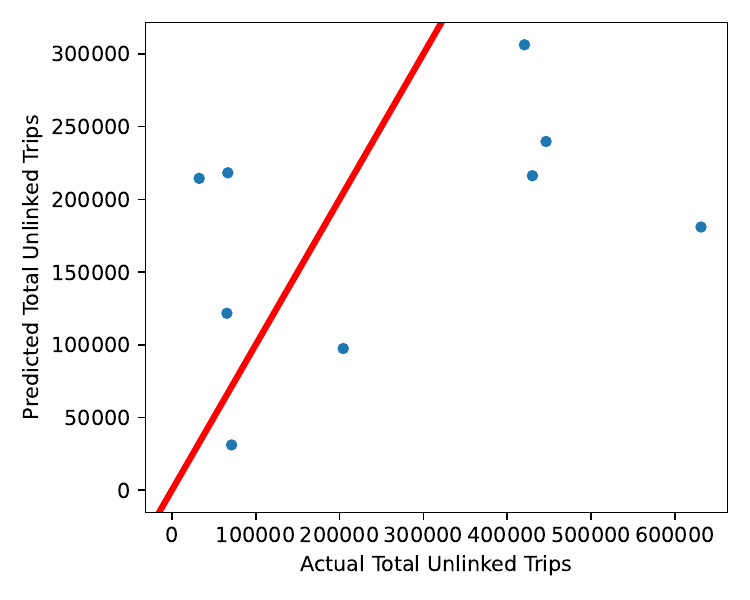}
    \caption{Each point represents the predicted demand versus the actual demand of the test dataset for a specific month, measured using the number of passenger trips. Points close to the red line of equation $y=x$ are the stops where prediction match real data.}
    \label{fig:temporal_demand_actual_vs_predicted}
\end{figure}

For demand, as illustrated in Table \ref{tab:demand_error}, a neural network again outperformed all other machine learning models. We employed a neural network with similar architecture as the one predicting supply.  In Figures \ref{fig:temporal_supply_actual_vs_predicted} and \ref{fig:temporal_demand_actual_vs_predicted}, the actual values are plotted against the values predicted by each model to examine the accuracy of the model. 

\begin{table}[b]
    \centering
    \begin{tabular}{c c c}
    \toprule
    Machine Learning Model & Root Mean Square Error & Relative Root Mean Square Error \\
    \midrule
    Linear Regression & 162946.267556 & 0.619120 \\
    Polynomial & 55475.328191 & 0.210780 \\
    Neural Network & 10624.695760 & 0.040369 \\
    Random Forest & 141924.549130 & 0.539247 \\
    \bottomrule
\end{tabular}
    \caption{Summary of performance measures of machine learning models for temporal demand, quantified as $NumberPassengerTrips$.}
    \label{tab:demand_error}
\end{table}

%

%
\subsection{Predicting Demand from Supply}

One of the main contributions of this paper is to provide the HDPT with a data-driven approach to predict demand from supply. Figures \ref{fig:linear_hours} and \ref{fig:linear_miles} reveal a linear relationship between monthly supply and demand, so we model this relationship using regular linear regression. The three outliers on both figures represent September, October, and November 2020. During this time, the COVID-19 pandemic was underway, where JMU was in and out of in-person classes for weeks until eventually classes went hybrid or completely online. During this time the buses were still running, leading to a supply surplus as observed from the graphs.

As the linear regression line in Figure \ref{fig:linear_hours} demonstrates, when vehicle revenue hours increase, so does passenger trips with a slope of 76.75. The linear regression line in Figure \ref{fig:linear_miles} also shows a positive slope: for every one mile increase in vehicle revenue miles per month, the HDPT can expect, on average, an increase in passenger trips for that month by approximately 10.54 trips. 

Overall, we can now predict the number of passenger trips from revenue miles and revenue hours using the following linear relationships: 

\begin{equation*}
    NumberPassengerTrips 
    = a_1 + b_1 (RevenueHours)
\end{equation*}

and 
\begin{equation*}
    NumberPassengerTrips 
    = a_2 + b_2 (RevenueMiles)
\end{equation*}

where $a_1,b_1,a_2,$ and $b_2$ are determined from the data. Figures \ref{fig:linear_hours} and \ref{fig:linear_miles} show the values of $a_1,b_1,a_2,$ and $b_2$ for the data that we incorporated in this study.

\begin{minipage}{0.47\textwidth}
    \includegraphics[width=\textwidth]{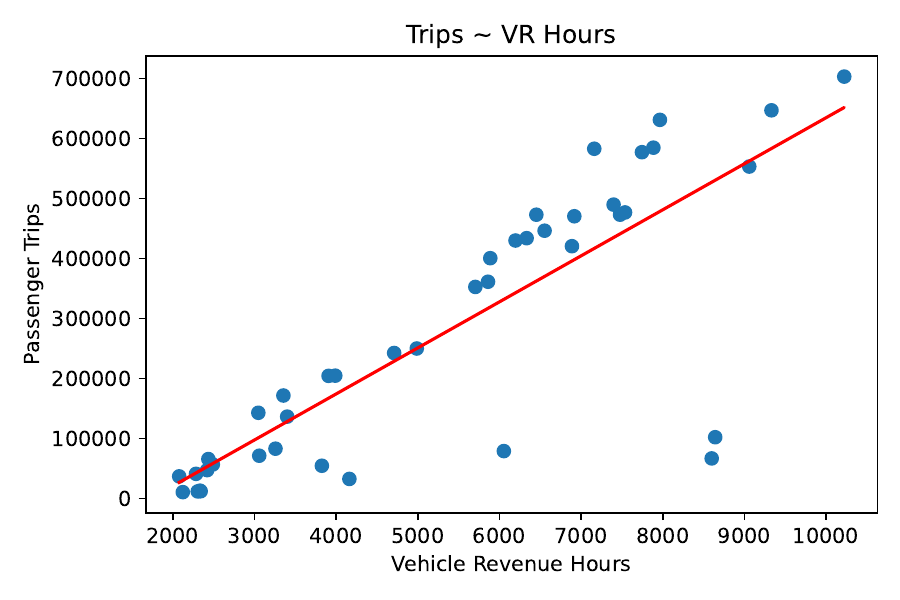}
    \captionof{figure}{A linear regression of $RevenueHours$ against $NumberPassengerTrips$.\\
    \\
    \footnotesize
    \emph{Trips = -184070.48 + 76.75(Hours)}}
    \label{fig:linear_hours}
\end{minipage}
\hfill
\begin{minipage}{0.47\textwidth}
    \includegraphics[width=\textwidth]{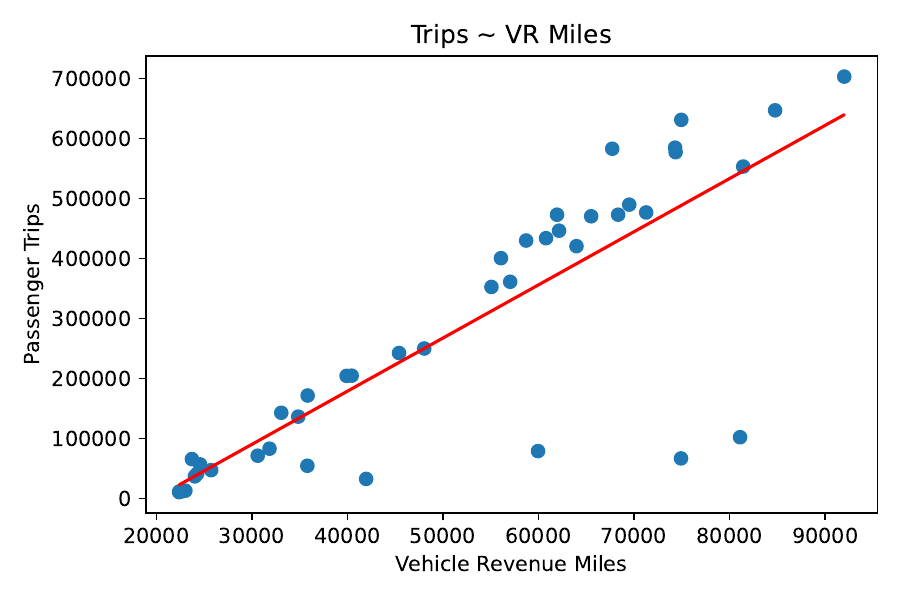}
    \captionof{figure}{A linear regression of $RevenueMiles$ against $NumberPassengerTrips$.
    \\
    \\
    \footnotesize
    \emph{Trips = -229380.05 + 10.54(Miles)}}
    \label{fig:linear_miles}
\end{minipage}
\subsection{Significant Predictors of Demand}

We investigate which variables are the most significant predictors of demand. This is important for the HDPT as it provides information on which factors warrant the most consideration when determining how their resources should be allocated. Moreover, it allows the HDPT to make necessary adjustments to their business model to meet the demand of their consumers. To this end, we define the significance of each input variable $x_n$ (representing median income, disability, vehicle ownership, and other input variables) as:

\begin{equation*}
    significance_{x_n}
    = \left\langle\frac{\partial Demand}{\partial x_n}\right\rangle.
\end{equation*}

In the above equation, we first compute the partial derivatives of the demand function with respect to the given variable $x_n$. We learned this nonlinear demand function using a neural network, so the partial derivative is not constant. It is given as a function of all input levels (discretized). To measure the average significance, we take the absolute value of the partial derivative at each input level then compute their average. This gives the average effect of the input variable on demand. 

\begin{figure}
    \centering
    \includegraphics[width=\textwidth]{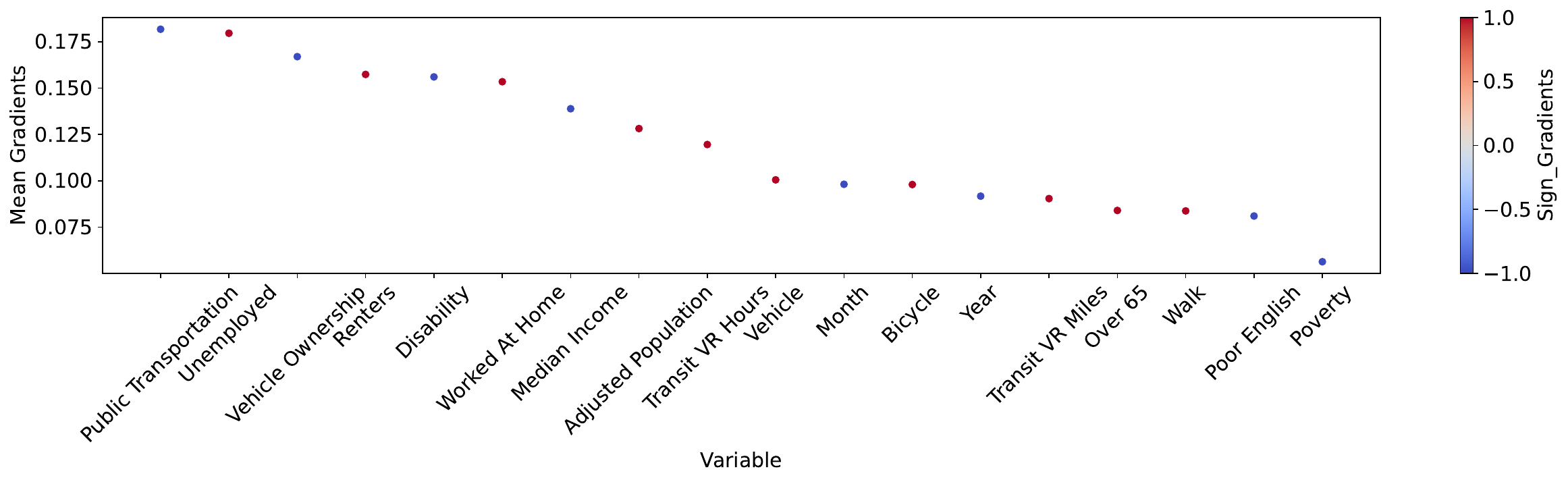}
    \caption{Significance of predictors of the temporal demand model are displayed. Red represents a positive average effect of the input variable on demand (if we only average the partial derivatives at each input level, without taking absolute values) while blue represents a negative average effect of the input variable on demand.}
    \label{fig:temporal_demand_sig}
\end{figure}

Figure \ref{fig:temporal_demand_sig} shows the results, identifying Public Transportation (the number of people who say they used public transportation as their means of travel), Unemployment, Vehicle Ownership, and Rental residence status as the most significant predictors for demand using this measure, within the framework of the temporal demand model. 

It is not surprising that self-reported use of public transportation as a favorite means of transportation, unemployment level, and vehicle ownership impact the demand for public transit. The significance of disability gives insight for the HDPT where there may be a gap in service to the community and that those with disabilities clearly demonstrate a demand for public transit services. However, Figure \ref{fig:temporal_demand_sig} also indicates that public transportation and disability have an inverse effect on ridership. This could be due to the fact that the census is self-reported, so more people will report they use public transportation to get to work because that is their ideal method, but in actuality they use other means. Due to the pandemic, less people might be reporting that they use the bus because of fear of the COVID-19 virus, especially those who are disabled and might also be considered high risk.

\section{Details of Spatial Supply and Demand Models} \label{sec:details_spatial_models}

\begin{figure}
    \centering
    \includegraphics[width=0.75\textwidth]{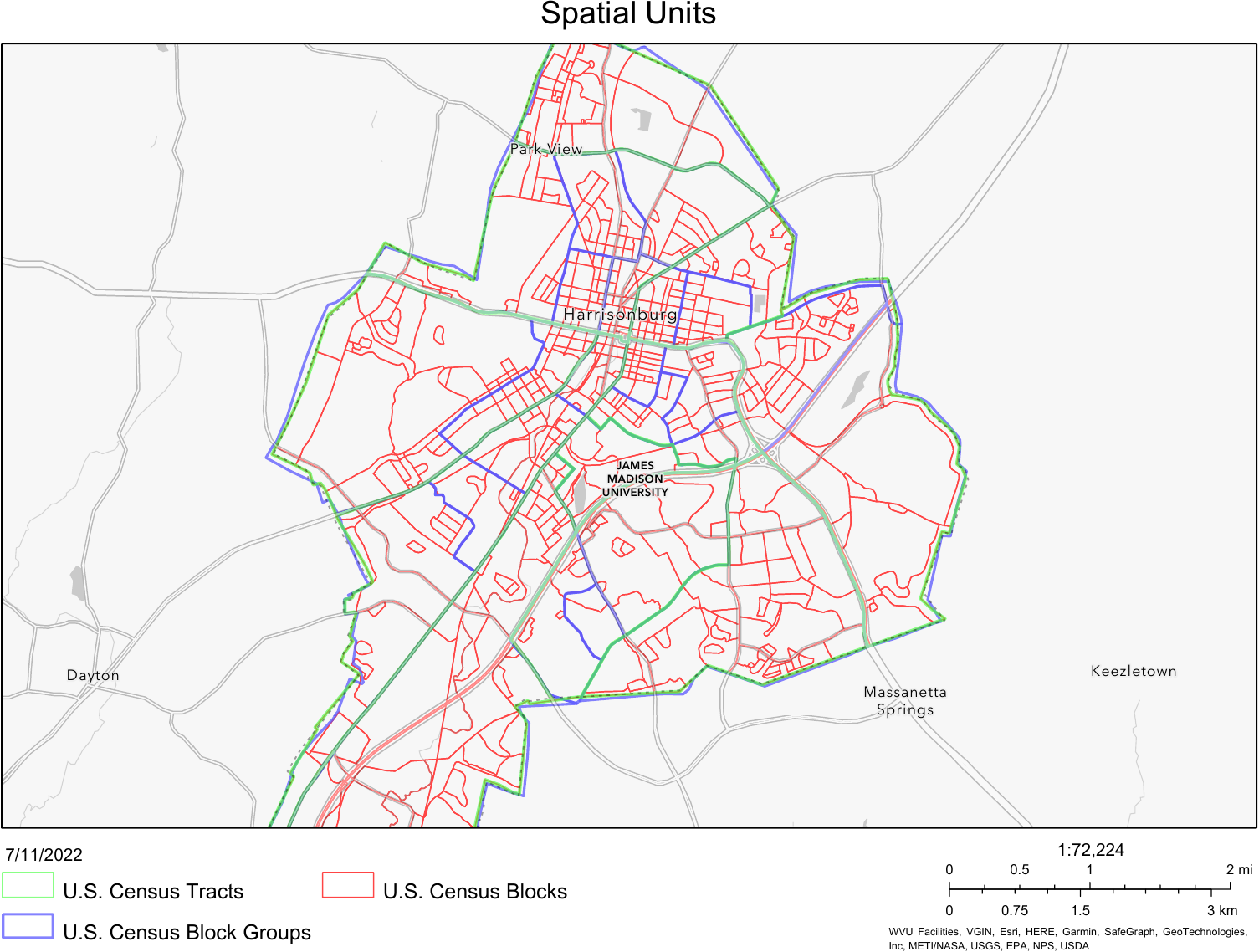}
    \caption{Visualization of U.S. Census Bureau Tracts, Blocks, and Block Groups in Harrisonburg.}
    \label{fig:census_spatial_units}
\end{figure}

While the temporal models are informative for the HDPT when considering monthly supply and demand, they do not provide any information on locational service gaps. To address these gaps, we construct models for supply and demand for each bus stop over the year 2019 (the year for which we had reliable data).  In this paper, we focus only on the stops serviced by the city bus routes, excluding all of the data related to the JMU bus routes.  This allowed us to study how the bus system served the permanent population of Harrisonburg and, in particular, its more socially vulnerable members.  While the U.S. Census Bureau provides a wealth of location-specific demographic information in Harrisonburg off of JMU's campus, broken down by block and block group as seen in Figure \ref{fig:census_spatial_units}, this granular level of data was not available for the people living on campus. Had the census data been available for the blocks surrounding JMU stops, then the same local analysis for supply and demand would apply to these stops as well.

\subsection{Data for Population Serviced by Each Stop} 
\begin{figure}
    \centering
    \includegraphics[width=0.5\textwidth]{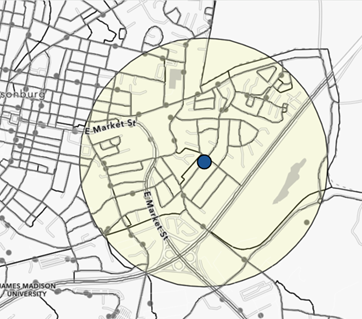}
    \caption{Example of three-quarter mile circle around a city bus stop for block service coverage.}
    \label{fig:stop_coverage}
\end{figure}

We start with the population density for each block in Harrisonburg city, together with the 2019 data from the U.S. Census Bureau, which is only available by block groups. To estimate the demographic characteristics for the population in each block, as opposed to a whole block group, we weigh each block group data by population density for each block in that group. We say that a bus stop services a block if any portion of the block is within a circle of radius three-quarters of a mile centered at the stop, as computed by ArcGIS and displayed in Figure \ref{fig:stop_coverage}. This is an actual circle as opposed to a walkable distance using sidewalks and walkable roads, which can be computed easily using ArcGIS. However, we chose to use the actual circle to remain consistent with current standards. Section 37.131 in the Americans with Disabilities Act of 1990 defines that anyone living within three-quarters of a mile from a bus stop is serviced by that stop and the HDPT utilizes this criteria for their routes and stops.

To estimate the characteristics for the population serviced by a particular stop, we combine the demographic values, weighted by population density, over all the blocks serviced by that stop. For example, we estimate a variable whose unit is $persons$, such as number of renters or the number of persons over the age of 65, first by multiplying the total population of each block by the percentage of the characteristic in the block group, and then summing over all of the blocks serviced by that stop.  For variables such as median income, we weigh the median income from each block group by the population density of the blocks serviced by a stop in that block group to calculate an average median income for the stop.  
\subsection{Quantifying Spatial Supply and Demand} 

In the spatial model, we quantify supply as $CityRoutesRan$, which is the total number of buses passing through a given stop in the city in the year 2019. Similar to the temporal model, we quantify demand at a given stop as the total ridership, denoted as $StopRidership$ (the target variable), which is the sum of the stop's boardings and alightings in 2019.  For this model we remove the three transfer hub data points.  These stops have very high recorded ridership values, for example, approximately 35\% of all ridership accounted for in our dataset was from the main transfer hub in Harrisonburg, but we cannot differentiate between passengers boarding/leaving the bus at these stops versus those transferring between different routes.   The ridership values at the transfer hubs are thus inflated due to these transfers, which is why we chose to exclude these stops from our demand model.   

\begin{figure}
    \centering
    \includegraphics[width=\textwidth]{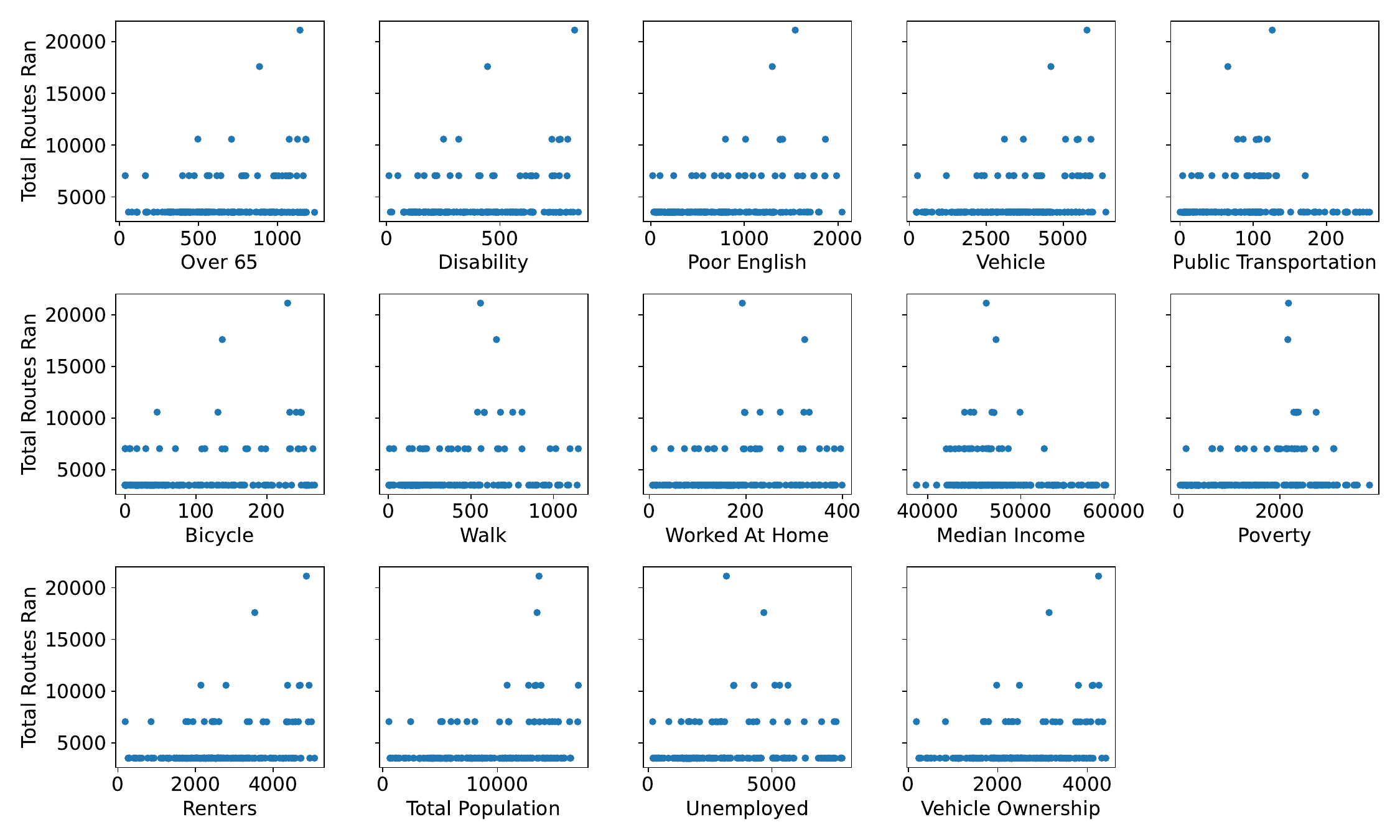}
    \caption{Scatter plots of $CityRoutesRan$ (quantifier for supply) against each social vulnerability input variable in the spatial supply model.}
    \label{fig:spatial_supply_vs_variables_plots}
\end{figure}

As for the predictor variables for the spatial models, we use the population serviced by each stop and the transit vulnerability characteristics corresponding to that population, denoted as $StopPop$ and $StopTVV$ (see Appendix \ref{table:spatial_variables} for the full list of those variables). As in the temporal model, $StopTVV$ is a compilation of multiple transit-related variables from the Social Vulnerability Index, given in Table \ref{tab:census_tract_svi}. This includes population age 65 and over, with disability, below poverty level, speak English ``less than well'', renter population, vehicle ownership, and means of transportation. These variables are important to include as they allow our models to take into account those who are vulnerable in Harrisonburg and, presumably, most in need of the transit system. Plots of each of the transit vulnerability variables (TVV) against our target supply variable, \textit{CityRoutesRan}, can be seen in Figure \ref{fig:spatial_supply_vs_variables_plots}. Figure \ref{fig:spatial_demand_vs_variables_plots} shows plots of each transit vulnerability variable against the target demand variable, \textit{StopRidership}.
\subsection{Spatial Supply Model}

We use the following function to predict spatial supply:
\begin{equation*}
    CityRoutesRan_{predicted} 
    = Supply \big(StopPop, StopTVV\big)
\end{equation*}

\noindent The above formula models how the characteristics of a population serviced by a given stop affect how often that stop is serviced. 

\subsection{Spatial Demand Model}

We use the following function to predict spatial demand, using as input variables the stop population data and the actual spatial supply at the stop:
\begin{equation*}
    StopRidership 
    = Demand \big(StopPop, StopTVV, CityRoutesRan_{actual}\big)  
\end{equation*} 

\begin{figure}
    \centering
    \includegraphics[width=\textwidth]{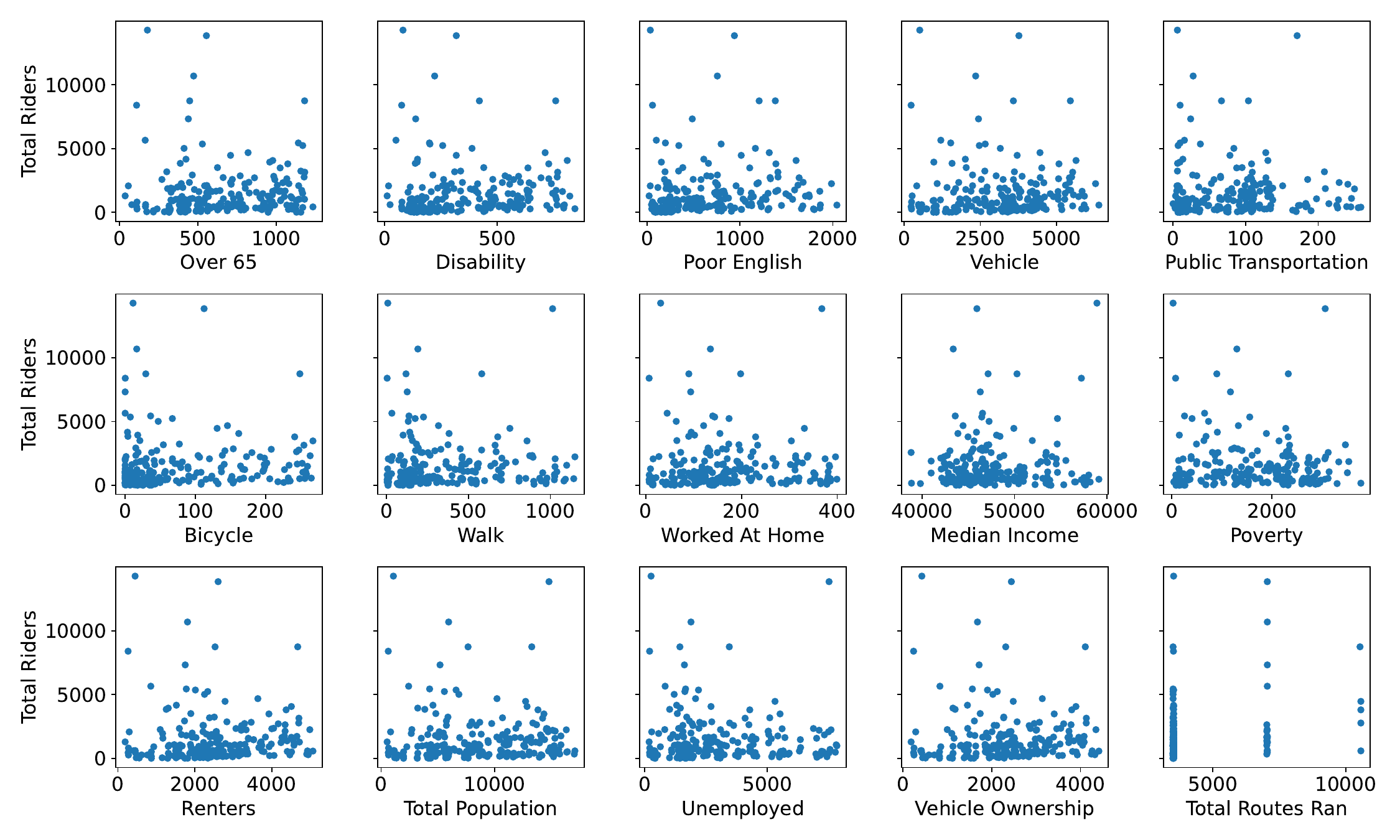}
    \caption{Scatter plots of $StopRidership$ (quantifier for demand) against each of the input variables in the spatial demand model.}
    \label{fig:spatial_demand_vs_variables_plots}
\end{figure}

Note that we used actual supply at a stop as input for building the demand model, but at prediction stage, or to experiment with changing levels of input, supply, or demand, we can use the predicted values of the spatial supply model values instead of the actual values, that is:

\begin{equation*}
    StopRidership = Demand \big(StopPop, StopTVV, CityRoutesRan_{predicted}\big) 
\end{equation*}

Therefore, we use the actual supply values for build stage, but the predicted supply values for deploy stage. 
\subsection{Machine Learning Models and Results}

Similar to temporal models, we experiment with various machine learning models: linear regression, random forests, and neural nets with different architectures. These models learn the supply and demand functions that best fit our data sets. Table \ref{tab:supply_spatial_error} and \ref{tab:demand_spatial_error} show the results from the models. Once again, neural nets perform the best in predicting spatial supply and demand, having the lowest relative root mean squared error. The neural networks we employed are feed forward fully connected neural networks with two hidden layers, ten nodes each, with ReLU activation functions.

\begin{table}
    \centering
    \begin{tabular}{c c c}
    \toprule
    Machine Learning Model & Root Mean Square Error & Relative Root Mean Square Error \\
    \midrule
    Linear Regression & 2866.742953 & 0.633664 \\
    Polynomial Regression & 19226.177394 & 4.249751 \\
    Neural Network & 171.666993 & 0.037945 \\
    Random Forest & 3562.326968 & 0.787416 \\
    \bottomrule
\end{tabular}
    \caption{Summary of performance measures of machine learning models for spatial supply, quantified as $CityRoutesRan$.}
    \label{tab:supply_spatial_error}
\end{table}

\begin{table}
    \centering
    \begin{tabular}{c c c}
    \toprule
    Machine Learning Model & Root Mean Square Error & Relative Root Mean Square Error \\
    \midrule
    Linear Regression & 1600.492575 & 1.399702 \\
    Polynomial Regression & 799915.791561 & 699.561962 \\
    Neural Network & 50.89870071 & 0.04451318 \\
    Random Forest & 3574.708237 & 3.126241 \\
    \bottomrule
\end{tabular}
    \caption{Summary of performance measures of machine learning models for spatial demand, quantified as $StopRidership$.}
     \label{tab:demand_spatial_error}
\end{table}

\begin{figure}
    \centering
    \includegraphics[width=0.45\textwidth]{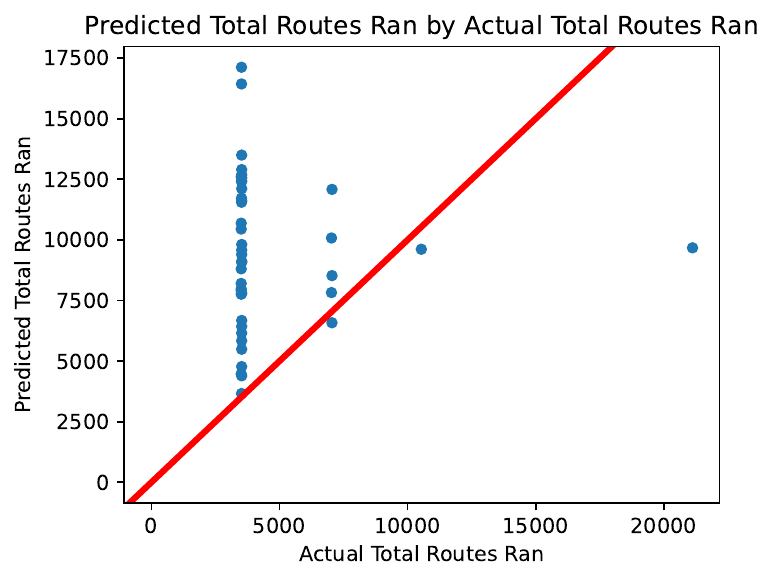}
    \caption{Each point represents the predicted supply versus the actual supply of the test dataset for a specific bus stop, measured using total routes ran. Points close to the red line of equation $y=x$ are the stops where the prediction matches the real data and stops far above or below the line correspond with under and over supplied stops, respectively.}
    \label{fig:spatial_supply_actual_vs_predicted}
\end{figure}

\begin{figure}
    \centering
    \includegraphics[width=0.45\textwidth]{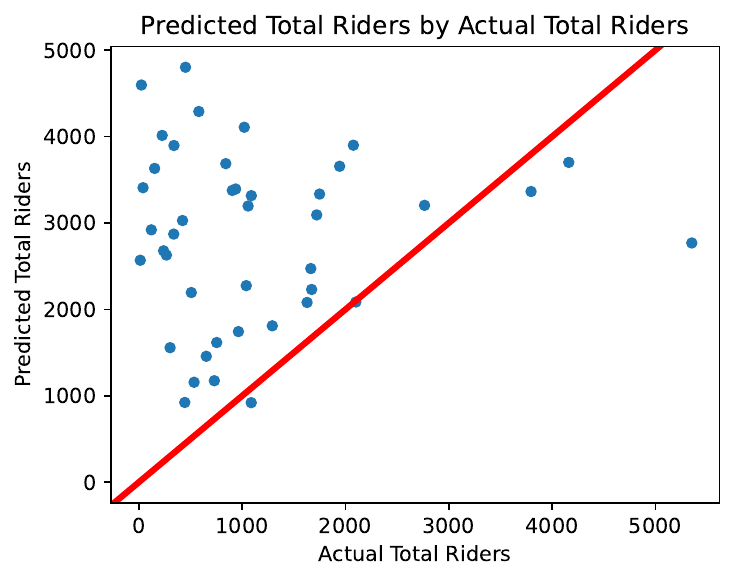}
    \caption{Each point represents the predicted demand vs the actual demand of the test dataset for a specific bus stop, measured using the number of passenger trips, with an outlier removed. Points close to the red line of equation $y=x$ are the stops where the prediction matches the real data and stops far above or below the line correspond with low and high demand stops, respectively.}
    \label{fig:spatial_demand_actual_vs_predicted}
\end{figure}

In Figures \ref{fig:spatial_supply_actual_vs_predicted} and \ref{fig:spatial_demand_actual_vs_predicted} the actual supply and demand values from the test data set are plotted against the values predicted by the respective models.  It is interesting to note that for the majority of the stops, the predicted values are higher, often significantly higher, than the actual values. The stop  on the far right of Figure \ref{fig:spatial_supply_actual_vs_predicted}, which has much higher actual supply versus predicted supply, is the main transfer hub.

\subsection{Assessing Service Gaps in Harrisonburg Using the Spatial Supply and Demand Models} 

\begin{figure}
    \centering
    \includegraphics[width=0.45\textwidth]{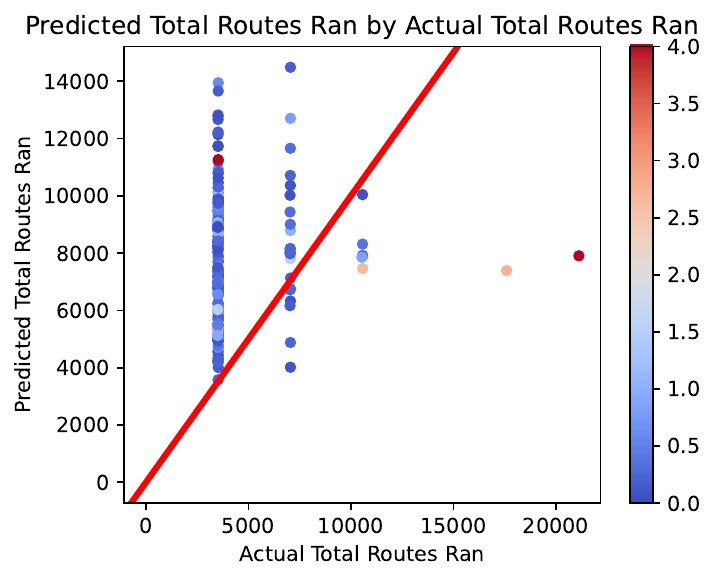}
    \caption{Each point represents the predicted supply versus the actual supply of the entire dataset for a specific bus stop, measured using $RevenueMiles$. Points close to the red line of equation $y=x$ are the stops where prediction match real data, and stops very far above or below the line correspond with under and over supplied stops respectively.  The color gradient represents the ratio of total ridership to total routes ran at each stop.}
    \label{fig:spatial_supply_actual_vs_predicted_entire}
\end{figure}

\begin{figure}
    \centering
    \includegraphics[width=0.45\textwidth]{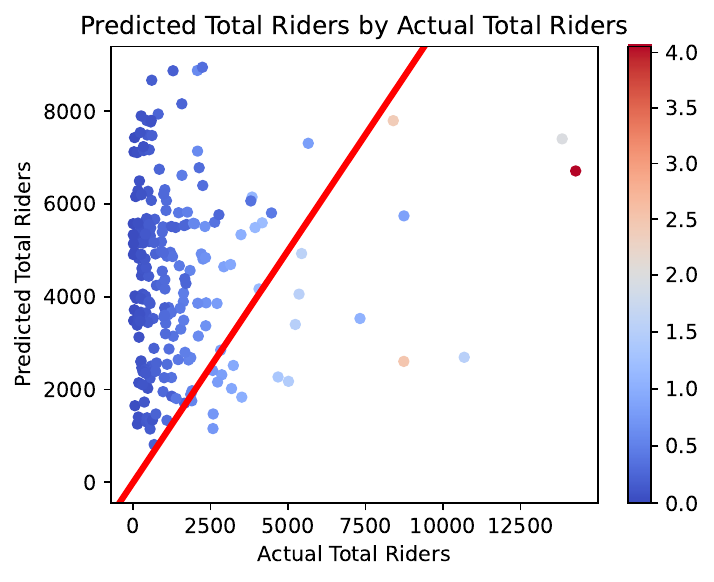}
    \caption{Each point represents the predicted demand vs the actual demand of the entire dataset for a specific bus stop, measured using the $NumberPassengerTrips$, with an outlier removed. Points close to the red line of equation $y=x$ are the stops where the prediction matches the real data and stops far above or below the line correspond with low and high demand stops, respectively.}
    \label{fig:spatial_demand_actual_vs_predicted_entire}
\end{figure}

In Figure \ref{fig:ridership_vs_routes_outlier} of subsection \ref{ssec:service_gap_real} we plotted the actual supply against the demand for each bus stop to assess service gaps for existing Harrisonburg city stops using real data on boarding, alighting, and number of routes passing through each stop (both in the training sets and test sets) in the city. Figures \ref{fig:spatial_supply_actual_vs_predicted_entire} and \ref{fig:spatial_demand_actual_vs_predicted_entire} plot actual vs predicted supply and demand values respectively for each stop; the color gradient reflects the ratio of total ridership to total routes ran (as in Figure \ref{fig:ridership_vs_routes_outlier}). While in subsection \ref{ssec:service_gap_real} we assessed gaps as locations with a high ratio of demand to supply (darker red spots), using the model we can identify potential gaps as stops where the predicted level of supply is significantly higher than actual level of supply (points in the upper left corner of Figure \ref{fig:spatial_supply_actual_vs_predicted_entire}). In comparing these two methods, we make the following observations:

\begin{itemize}
    \item The stops with a high ratio of demand to supply on the right side of Figure \ref{fig:spatial_supply_actual_vs_predicted_entire} are each transfer hubs which significantly skews both the actual supply and demand values and are not accounted for in the models.  
    
    \item Of the stops in the leftmost column of Figure \ref{fig:spatial_supply_actual_vs_predicted_entire}, these are stops that lie on a single route and are serviced once per loop. The model predicts the one with the highest ratio of demand to supply to be serviced much more often than it currently is.  This stop lies in a heavily populated area in the northwestern edge of the city.  The next two highest ratio stops in this column do not lie in residential areas, one services Harrisonburg High School while the other is outside the DMV offices, and their predicted supply is significantly lower.
\end{itemize}

These predicted values are reflective of the demographic information of the residents around each stop  that is primarily incorporated into these models; this comparison suggests that we can potentially improve the models by including additional information about the businesses, schools, and government offices around each stop. 

\subsection{Significant Predictors of Spatial Demand}

\begin{figure}
    \centering
    \includegraphics[width=\textwidth]{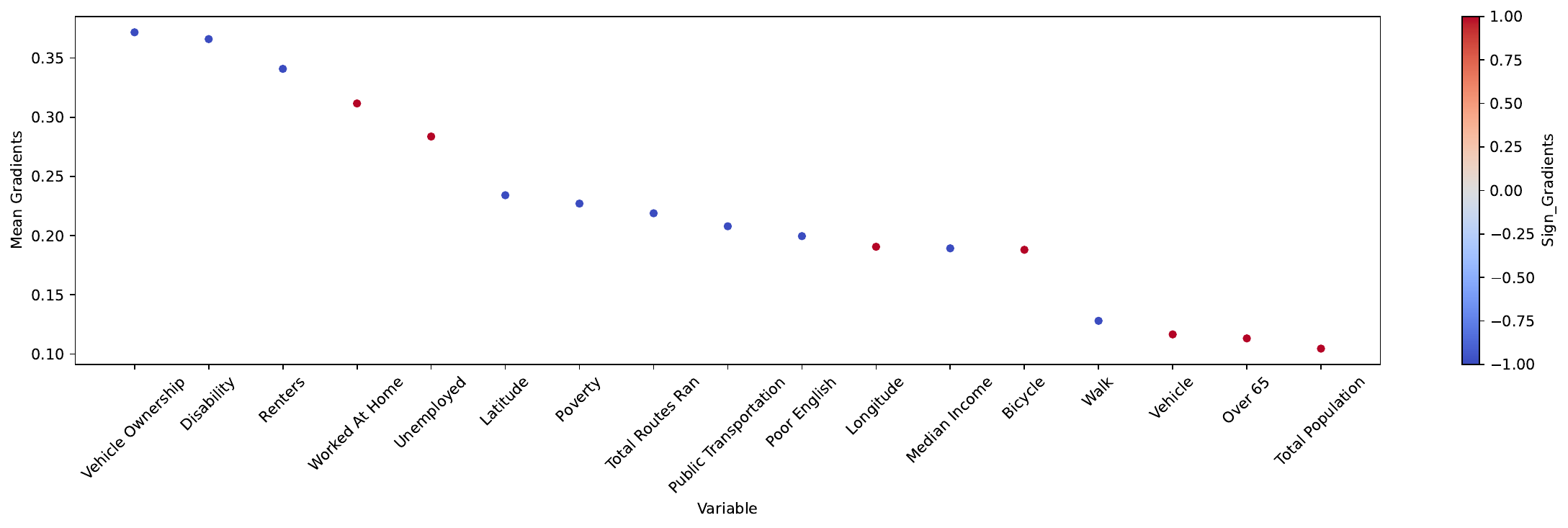}
    \caption{Significance of predictors of the spatial demand model. Red represents a positive average effect of the input variable on demand (if we only average the partial derivatives at each input level, without taking absolute values) while blue represents a negative average effect of the input variable on demand.}
    \label{fig:spatial_demand_sig}
\end{figure}

To determine the significant predictors of stop ridership, we calculated the partial derivative of the spatially dependent demand function with respect to each of its input variables, as done in the temporal model. Figure \ref{fig:spatial_demand_sig} shows the results, identifying Vehicle Ownership, Disability, and Rental residence status as the most significant predictors for demand using this measure, and within the framework of the spatial demand model.

\section{Conclusions, Recommendations, and Future Work} \label{sec:discussion}

In this paper, we assess the supply and demand for the bus system in Harrisonburg city using two data-driven models, one temporal, and one spatial. Our models take into account the city's population distribution and attributes, social vulnerability, and the significant changes in population when JMU is on or off session. Based on this study, we identify service gaps, pinpoint unserviced areas, and determine significant predictors for supply and demand. The following subsections summarize our recommendations for similar studies that attempt to assess and enhance the performance and efficiency of transportation systems. 
\subsection{Addressing Gaps in Supply}

The temporal models demonstrated a linear relationship between supply and demand, as shown in Figures \ref{fig:linear_hours} and \ref{fig:linear_miles}. This suggests that the HDPT does a good job of adjusting the overall supply of buses in response to changes in demand on a month to month basis. This is them mostly accounting for when JMU is or is not in session.

As for the spatial model, there is little correlation between the ridership of an individual stop and number of times a bus services the stop (see Figure \ref{fig:spatial_demand_vs_variables_plots} in the bottom right).  Outside of transfer hubs and stops near transfer hubs that are serviced by multiple routes, Figures \ref{fig:spatial_supply_actual_vs_predicted_entire} and \ref{fig:spatial_demand_actual_vs_predicted_entire} show that the models predict much higher supply and demand for  the majority of individual stops. This, in combination with our conclusions from subsection  \ref{ssec:model_overview} that many of the higher demand stops are on the outer edges of the city suggests that the current routes and schedules are not optimizing efficiency. Rather than having one main transfer hub in the downtown area, the increased  demand for buses on the outskirts of the city suggests that a decentralized route system may be better in terms of efficiently allocating resources.  
\subsection{Data Quality and Data Collection Recommendations}

Our world has become heavily reliant on data to make decisions. It is therefore vital to invest in efficient and effective methods for the collection and storage of data required for proper and informed decision making. During this project, we encountered multiple obstacles due to data quality issues. 

\begin{itemize}
    \item It is useful to collect data continuously and by various time intervals---daily, weekly, monthly, and annually. We found that, for our purposes, the most common data collection time interval was annual data. While this data is useful to some extent, making the assumption that the data aspects remained constant over the years is less preferable compared to a more granular approach with smaller units of time.
    
    \item It is important to accompany data sets with the transformations and mathematical formulas that produced the given values in the data tables. For instance, the social vulnerability index (SVI) is an important part of our project; however, there is no formula that is easily accessible or well-explained by the CDC that we could use with our own variables to calculate the SVI for the city blocks during a given year. We therefore had to include raw variables in our models, denoted as $TVV$. It would be beneficial to make this formula---and other commonly used formulas---readily available so that researchers can have this essential information at their disposal.
    
    \item It is essential that organizations keep raw data records rather than only pulling summary reports when data is requested. Having access to raw data is highly beneficial to data analysts. We find it likely that our data was corrupted when it was compiled as a report. This issue could have been avoided were we given access to raw data files. Therefore, a good starting point for any company looking to optimize their data collection would be the proper storage of their data as raw data files.
\end{itemize}
\subsection{Accounting For More Variables} 

When data becomes available, we can account for more variables in our temporal and/or spatial supply and demand models. Ideally, we would like to include:

\begin{itemize}
    \item the type of area within each bus stop's locality: residential, school, commercial, industrial, etc. Each locality will have a percentage of every type (including zero percent or one hundred percent);
    \item the number of bus drivers available within a given time period;
    \item the number of transit vehicles in operation within a given time period;
    \item the type of technology that vehicles are equipped with;
    \item the overall service coverage;
    \item shared rides usage in the area (such as Uber and Lyft);
    \item scooter usage in the area (such as Bird or Lime);
    \item bus stop visibility;
    \item bus mobile app availability and usage.
\end{itemize}
\subsection{Using Real Walkable Distance from the Bus Stop} 

A bus stop services a block if any portion of the block is within a circle of radius three-quarters of a mile centered at the stop, as computed by ArcGIS. The circle is currently calculated as the crow flies rather than using the walkable distance over sidewalks and roads. This walkable distance can also be computed using ArcGIS, but we chose to use distance as the crow flies to be in accordance with the standards currently in Section 37.131 of the Americans with Disabilities Act of 1990. This act defines anyone living within three-quarters of a mile from a bus stop as serviced by that stop, so this is the current definition used by the HDPT. Given that the current GIS technology can account for actual walkable distances, we recommend that the standards be updated accordingly. 
\subsection{Future Work} 

Due to a data scramble, we were unable to analyze patterns and build models across multiple years. This influenced our decision to construct a spatial model instead of a spatiotemporal model. Ideally, building a spatiotemporal model would allow analysis across both a space and a time range with the proper data to support the construction of these models.

\subsubsection{Building upon Service Gaps}

To better assess service gaps, we hope to rerun our wrangling and modeling on five years of stop data that has the correct ridership for each stop. The scrambling of the data limited the time period analyzed by this project; however, if we have access to raw data files, it would be possible to utilize our models over longer time frames.

\subsubsection{Assessing Efficiency}

We would like to continue towards a model which would demonstrate the amount of time it would take to travel from Point A to Point B to quantify the efficiency of the transit system. Using the current bus schedules, the travel time between all stops and access places with high times would be mapped out. The target variable would be the time it takes to travel between any two given locations on any given day. This would take place by utilizing graph theory, where the stops are nodes and the distances between stops are the edges.

\subsubsection{Assessing Connectivity}

The connectivity of the bus system and the availability of information of potential riders must also be addressed, as this could also give insights on accessibility. Additionally, the HDPT intends to hire more full-time drivers, so with increased resources comes the opportunity to optimize allocation and reap the benefits of their labor.

\section*{Author Contribution Statement}
Authors Miranda Bihler, Hala Nelson, Erin Okey, Noe Reyes Rivas, John Webb, and Anna White contributed to the design and implementation of the research, to the analysis of the
results and to the writing of the manuscript.

\section*{Acknowledgments}
This work was supported by the National Science Foundation (NSF) grant 1950370 and by James Madison University's College of Science and Mathematics. Any opinions, findings, and conclusions or recommendations expressed in this material are those of the authors and do not necessarily reflect the views of the National Science Foundation. 

\appendix
\section{Table of Variables for the Temporal Supply and Demand Models} \label{tab:supply_variables}

\renewcommand{\arraystretch}{1}
\begin{longtable}{>{\raggedright}p{3cm} >{\raggedright}p{2.5cm} >{\raggedright}p{4.5cm}  p{5cm}}
    \toprule
     Variable & Model & Source & Variable Definition and Construction \\
   \midrule
   \endhead
    
     \endfoot
     \endlastfoot
     Year & Supply and Demand & HDPT & Time value \\
     Month & Supply and Demand & HDPT & Time value \\
    Adjusted population & Supply & \acs \ \& JMU Office of Institutional Research & Population in Harrisonburg, VA when accounting for the change in population in regards to the JMU academic calendar \\
    Population age 65 and over & Demand & \acs & Amount of residents in Harrisonburg, VA with an age of 65 and older \\
    With disability & Demand & \acs & Income level of a given household where half the households earn more and half earn less than this value in a specific area of interest \cite{uscb2020american} \\
    Below poverty level & Demand & \acs & A family's total income (income before taxes and not including capital gains or noncash benefits) is less than the family's threshold, with poverty threshold determined by the U.S. Census Bureau \cite{uscb2020american} \\
    Speak English ``less than well'' & Demand & \acs & Based on survey respondents' self-perception on their English speaking abilities \cite{uscb2020american} \\
    JMU Enrollment & Supply & JMU Office of Institutional Research & All students enrolled in a degree program at JMU \\
    JMU routes ran & Supply & HDPT & Count of how many buses passed through a certain stop on the JMU routes on a given day \\
    City routes ran & Supply & HDPT & Count of how many buses passed through a certain stop on the city routes on a given day \\
    Means of transportation (private vehicle, public transit, bicycle, walking, worked at home) & Demand & \acs & Principal mode of travel or type of conveyance that the worker usually used to get from home to work during the reference week, separated by the type of transportation \cite{uscb2020american} \\
    Vehicle Ownership & Demand & \acs & Number of passenger cars, vans, and pickup or panel trucks of one-ton (2,000 pounds) capacity or less kept at home and available for the personal use of household members \cite{uscb2020american} \\
    Median income & Demand & \acs & Value that falls in the middle of all household incomes that are received on a regular basis before payments for taxes, social security, etc. and does not reflect noncash benefits \\
    Renter population & Demand & \acs & Number of people that do not own their dwelling \cite{uscb2020american} \\
    Adjusted Population & Demand & \acs{} \& Office of Institutional Research JMU & Population in Harrisonburg, VA when accounting for the change in population based on JMU academic calendar \\
    Unemployed population & Demand & \acs & All civilians 16 years old and over are classified as unemployed if they (1) were neither “at work” nor “with a job but not at work” during the reference week, and (2) were actively looking for work during the last 4 weeks, and (3) were available to start a job \cite{uscb2020american} \\
    Transit vehicle revenue miles & Demand & HDPT & The miles that vehicles are scheduled to or actually travel while in revenue service \\
    Transit vehicle revenue hours & Demand & HDPT & The hours that vehicles are scheduled to or actually travel while in revenue service \\
    \bottomrule
\end{longtable}

\newpage
\section{Table of variables for the spatial supply and demand models} \label{table:spatial_variables}
\begin{longtable}{>{\raggedright}p{3cm} >{\raggedright}p{2.5cm} >{\raggedright}p{4.5cm} p{5cm}}
    \toprule
    Variable & Model & Source & Variable Definition and Construction \\
    \midrule
    \endhead  
    \endfoot
    \endlastfoot

    Latitude & Supply and Demand & ArcGIS & Location of bus stops \\
    Longitude & Supply and Demand & ArcGIS & Location of bus stops \\
    Total Population & Supply and Demand & \acs & Population in Harrisonburg, VA \\
    Total routes ran per day & Demand & HDPT & Calculated by number of routes in service on a given day, assuming one bus is used per route \\
    With disability & Supply and Demand & \acs & Someone who reports having serious difficulty with specific functions---hearing, vision, cognition, and ambulation---and may, in the absence of accommodation, have a disability \cite{uscb2020american} \\
    Speak English ``less than well'' & Supply and Demand & \acs & Based on survey respondents' self-perception on their English speaking abilities \cite{uscb2020american} \\
    Vehicle ownership & Supply and Demand & \acs & Number of passenger cars, vans, and pickup or panel trucks of one-ton (2,000 pounds) capacity or less kept at home and available for the personal use of household members \cite{uscb2020american} \\
    Population age 65 and over & Supply and Demand & \acs & Amount of residents in Harrisonburg, VA with an age of 65 and older \\
    Median income & Supply and Demand & \acs & Value that falls in the middle of all household incomes that are received on a regular basis before payments for taxes, social security, etc. and does not reflect noncash benefits \\
    Below poverty level & Demand & \acs & A family's total income (income before taxes and not including capital gains or noncash benefits) is less than the family's threshold, with poverty threshold determined by the U.S. Census Bureau \cite{uscb2020american} \\
    Renter population & Supply and Demand & \acs & Number of people that do not own their dwelling \cite{uscb2020american} \\
    Unemployed population & Supply and Demand & \acs & All civilians 16 years old and over are classified as unemployed if they (1) were neither ``at work'' nor ``with a job but not at work'' during the reference week, and (2) were actively looking for work during the last 4 weeks, and (3) were available to start a job \cite{uscb2020american} \\
    Means of transportation (private vehicle, public transit, bicycle, walking, worked at home) & Supply and Demand & \acs & Principal mode of travel or type of conveyance that the worker usually used to get from home to work during the reference week, separated by the type of transport \\
    \bottomrule
\end{longtable}

\clearpage
\printbibliography
\end{document}